\documentclass{article} 
\usepackage{iclr2025_conference,times}

\usepackage{amsmath,amsfonts,bm}

\def\eqref#1{equation~\ref{#1}}

\def\1{\bm{1}}

\DeclareMathAlphabet{\mathsfit}{\encodingdefault}{\sfdefault}{m}{sl}
\SetMathAlphabet{\mathsfit}{bold}{\encodingdefault}{\sfdefault}{bx}{n}

\usepackage{hyperref}
\usepackage{graphicx}
\usepackage{url}
\usepackage{calligra}
\usepackage{amsmath}
\usepackage{listings}
\usepackage{booktabs}
\usepackage{multirow}
\usepackage{wrapfig}
\usepackage{makecell}
\usepackage{listings}
\usepackage{xcolor}
\usepackage{appendix}
\newcommand{\T}[1]{\texttt{#1}}
\newcommand{\logicalOR}{\; | \;}

\definecolor{codegreen}{rgb}{0,0.6,0}
\definecolor{codegray}{rgb}{0.5,0.5,0.5}
\definecolor{codepurple}{rgb}{0.58,0,0.82}
\definecolor{backcolour}{rgb}{0.95,0.95,0.92}

\lstdefinestyle{mystyle}{
    commentstyle=\itshape\color{codegreen}, 
    keywordstyle=\bfseries\color{magenta}, 
    stringstyle=\color{codepurple}, 
    basicstyle=\ttfamily\tiny,
    breakatwhitespace=false, 
    breaklines=true, 
    keepspaces=true, 
    showspaces=false, 
    showstringspaces=false, 
    showtabs=false, 
    tabsize=1, 
}

\title{Unveiling the Magic of Code Reasoning \\ through Reflective Hypothesis \\ Decomposition and Amendment}

\author{Yuze Zhao$^1$, Tianyun Ji$^{1,*}$, Wenjun Feng$^{1,*}$, Zhenya Huang$^{1,2,\dag}$, Qi Liu$^{1,2}$, \\ {\bf Zhiding Liu$^{1}$, Yixiao Ma$^{1}$, Kai Zhang$^{1}$, Enhong Chen$^{1}$} \\
$^1$State Key Laboratory of Cognitive Intelligence,\\
University of Science and Technology of China\\
$^2$Institute of Artificial Intelligence, Hefei Comprehensive National Science Center \\
\texttt{yuzezhao@mail.ustc.edu.cn huangzhy@ustc.edu.cn}}

\iclrfinalcopy 
\begin{document}

\maketitle

\def\thefootnote{*}\footnotetext{Equal contribution}\def\thefootnote{\arabic{footnote}}

\def\thefootnote{\dag}\footnotetext{Corresponding author}\def\thefootnote{\arabic{footnote}}

\begin{abstract}
The reasoning abilities are one of the most enigmatic and captivating aspects of large language models (LLMs). Numerous studies are dedicated to exploring and expanding the boundaries of this reasoning capability. However, tasks that embody both reasoning and recall characteristics are often overlooked. In this paper, we introduce such a novel task, \textbf{code reasoning}, to provide a new perspective for the reasoning abilities of LLMs.
We summarize three meta-benchmarks based on established forms of logical reasoning, and instantiate these into eight specific benchmark tasks. Our testing on these benchmarks reveals that LLMs continue to struggle with identifying satisfactory reasoning pathways.
Additionally, we present a new pathway exploration pipeline inspired by human intricate problem-solving methods. This \textbf{R}eflective \textbf{H}ypothesis \textbf{D}ecomposition and \textbf{A}mendment (\textbf{RHDA}) pipeline consists of the following iterative steps: (1) Proposing potential hypotheses based on observations and decomposing them; (2) Utilizing tools to validate hypotheses and reflection outcomes; (3) Revising hypothesis in light of observations. Our approach effectively mitigates logical chain collapses arising from forgetting or hallucination issues in multi-step reasoning, resulting in performance gains of up to $3\times$. Finally, we expand this pipeline by applying it to simulate complex household tasks in real-world scenarios, specifically in VirtualHome, enhancing the handling of failure cases. We release our code and all of results at \url{https://github.com/TnTWoW/code_reasoning}.
\end{abstract}

\section{Introduction}
Large Language Models (LLMs), which are trained on billions of tokens, have demonstrated impressive reasoning abilities in complex tasks~\citep{brown2020language,wei2022chain,kojima2022large,openai2023gpt4}. 
However, it is evident that as potential fuzzy retrieval systems or parameterized knowledge compression systems~\citep{xie2021explanation}, LLMs perform better on System 1 tasks than on System 2 tasks~\citep{kahneman2011thinking, Bengio2019from, yao2023tree, weston20232attention, liu2023guiding}. Specifically, LLMs excel in intuitive memory retrieval tasks, but continue to face significant challenges with tasks requiring rational reasoning~\citep{kambhampati2024can}.

From the perspective of human cognitive psychology, \textbf{reasoning can be viewed as a process of memory retrieval}, in which people retrieve relevant information from memory and use it to make inferences~\citep{Kyllonen1990Reasoning,SU2002Working,hayes2014memory,feeney2014reasoning,Kyle2015Reasoning}. For example, \citet{haidt2001emotional} proposed that when individuals engage in moral reasoning, they typically draw on their prior knowledge from social and cultural contexts.
Similarly, studies involving animal lesions and human neuroimaging have confirmed that the hippocampus, which is primarily associated with memory, also plays a crucial role in reasoning abilities~\citep{zeithamova2012hippocampus}.
Therefore, memory and reasoning are interdependent, with considerable overlap between the two, rendering the distinction between them somewhat arbitrary~\citep{Heit2012Relations, liu2023learning}.

\begin{figure}[t]
    \centering
    \includegraphics[width=\linewidth]{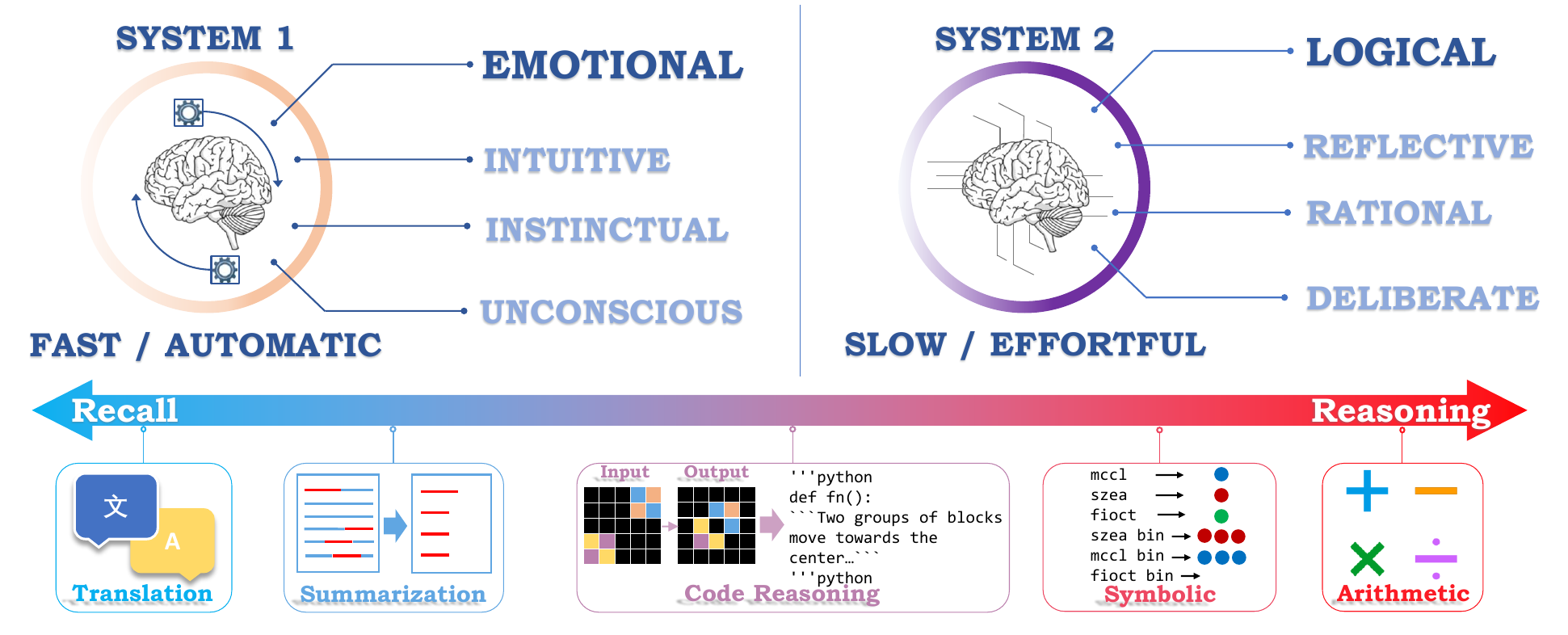}
    \caption{Code reasoning is a category of tasks that incorporates logical reasoning into code, aiming to solve programming problems through logical reasoning. These tasks require a balance between background knowledge and thinking span, placing greater emphasis on the collaborative functioning of both System 1 and System 2 thinking.}
    \label{fig:intro}
    \vspace{-5mm}
\end{figure}
 
We believe that, similar to humans~\citep{strachan2024testing, liu2024socraticlm, lin2024learning}, LLMs do not exhibit a clear boundary between memory and reasoning~\citep{schaeffer2024emergent, razeghi2022impact}. However, tasks that lie at this intersection are often overlooked in research. Here, we propose a novel task to explore the capability boundaries of LLMs: \textbf{Code Reasoning}. Code reasoning encompasses a category of tasks that demonstrate logical reasoning through code and address problems in a systematic manner.
As illustrated in Figure~\ref{fig:intro}, we position some tasks along an axis that reflects 1) the degree of reliance on prior knowledge (Recall) and 2) the extent to which prior knowledge is applied to the current context (Reasoning). We position the code reasoning task between memory and reasoning.  On one hand, the highly structured nature of code requires the model to learn syntax from pre-training data, enabling it to recall relevant information during solving a problem. 
On the other hand, generating code solutions necessitates the model's understanding of the problem and context, involving reasoning to produce appropriate solutions. 

In this paper, we introduce code reasoning, a task that formalizes reasoning steps into a programming language and offloads the computation process to the compiler. To explore different aspects of code reasoning, we summarize three meta-benchmarks based on existing forms of logical reasoning: inductive code reasoning, deductive code reasoning, and abductive code reasoning.

Inductive code reasoning involves deriving broad generalizations from a series of observations, demonstrating the ability to infer rules from examples and generate programs to meet input-output mapping. Deductive code reasoning starts from premises and derives valid conclusions, focusing on the model's capacity to understand a program's intermediate states and reasoning step by step. Abductive code reasoning seeks the simplest and most likely explanation based on a set of observations, highlighting the model's ability to abstractly understand a function's purpose. 

We concretize these three meta-benchmarks into eight specific benchmarks. Based on these eight benchmarks, we evaluate the performance of existing models in code reasoning. Due to data sparsity, we find that current state-of-the-art LLMs still struggle to achieve satisfactory results in solving such problems. To enhance the reasoning process, we implement a \textbf{R}eflective \textbf{H}ypothesis \textbf{D}ecomposition and \textbf{A}mendment (\textbf{RHDA}) pipeline.
This pipeline is iterative, encompassing hypothesis decomposition, execution verification, and amendment submission.
Specifically, we first guide the LLM to formulate initial hypotheses based on complex observations and decompose these into sub-hypotheses. These sub-hypotheses are then compiled into executable functions through a translator, enabling direct application to the observations, followed by validation using external tools. Subsequently, based on the execution results and observations, the LLM submits amendments to reflect on and refine the issues within the sub-hypotheses.

Our experimental results indicate that RHDA methods effectively mitigate reasoning failures caused by data sparsity. With the same or even lower overhead, this method achieved performance improvements of up to three times compared to baseline methods. Finally, we extend this pipeline to complex, simulated real-world household tasks VirtualHome~\citep{puig2018virtualhome, puig2020watchandhelp}, guiding the LLM to complete a series of intricate operations.

\section{Meta-Benchmark}
We describe the general process of code reasoning as the transformation from Input $\mathcal{I}$ and Program $\mathcal{P}$ to Output $\mathcal{O}$, represented as $\mathcal{I}\stackrel{\mathcal{P}}{\longrightarrow}\mathcal{O}$. Inductive code reasoning is concretized as the Programming by Example (PBE) task. In this task, a neural program synthesis model $\mathcal{M}$ searches the execution space to find a program that best satisfies all given input-output specifications. We donate this meta-benchmark as $\mathcal{M}(\mathcal{I}, \mathcal{O})\rightarrow\mathcal{\widetilde{P}}$.
Deductive code reasoning is exemplified in tasks that simulate the program execution process. In this task, a neural simulation compiler model $\mathcal{M}$ tracks the program's execution and records intermediate states, gradually deriving the final valid output. We denote this meta-benchmark as $\mathcal{M}(\mathcal{I}, \mathcal{P})\rightarrow\mathcal{\widetilde{O}}$.
Abductive code reasoning is concretized as input prediction tasks. This task requires the neural understanding model $\mathcal{M}$ to form an abstract-level understanding of function's behavior and perform abductive inference based on the given program and output. We represent this meta-benchmark as $\mathcal{M}(\mathcal{O}, \mathcal{P})\rightarrow\mathcal{\widetilde{I}}$.
The details of the benchmarks are provided in the Appendix~\ref{app:benchmark_details}.

\subsection{Inductive Code Reasoning}
Inductive code reasoning can be represented as $\mathcal{M}(\mathcal{I}, \mathcal{O})\rightarrow\mathcal{\widetilde{P}}$ and is concretized as a PBE task~\citep{qiu2024phenomenal, shi2024exedec}. PBE is a program synthesis task designed to help end-users, particularly non-programmers, create scripts to automate repetitive tasks~\citep{gulwani2016programming}. Based on input-output specifications, PBE systems can synthesize a program in either a general-purpose language (GPL) or a domain-specific language (DSL). 
Inductive code reasoning encompasses four challenging PBE tasks, two of which are GPL tasks: List Function~\citep{rule2020child} and MiniARC~\citep{kim2022playgrounds}, while the other two are DSL tasks: RobustFill~\citep{devlin2017robustfill} and DeepCoder~\citep{balog2016deepcoder}.
GPL tasks are relatively complex, allowing the model to solve problems in a more flexible manner. In contrast, DSL tasks require the model to quickly learn the syntax of DSL through few-shot learning and address relatively simpler problems.

\paragraph{List Function.} The List Function task was originally designed to investigate how humans learn the concept of computable functions that map lists to lists. Given input and output specifications in the form of lists, the model generates GPL rules that conform to these specifications. For example, with an input specification of \texttt{[2, 4, 8, 10]} and an output specification of \texttt{[3, 5, 9, 11]}, we expect the resulting rule to be \texttt{lambda x : x + 1}\footnote{For conciseness while maintaining generality, we will use lambda expressions to represent a program.}.

\paragraph{MiniARC.} MiniARC is a compressed 5x5 version of the Abstraction and Reasoning Corpus~\citep{chollet2019measure, moskvichev2023concept}, designed to assess imaginative and reasoning abilities.
MiniARC balances the length of the input-output pairs with the difficulty of the problems. The specifications are 5x5 2D grids, where the numbers represent blocks of specific colors. The model must find valid problem-solving paths (such as color swapping, row flipping) to achieve the transformation from input to output.

\paragraph{RobustFill.} RobustFill is a string manipulation task where the model is expected to perform a combination of atomic operations, such as extracting a substring from position $k_1$ to $k_2$ using $SubString(k_1, k_1)$, to achieve generalization.
As an example, a program \texttt{ToCase(Lower, SubStr(1,3))} converts full month names (January, April) to their abbreviations (jan, apr).

\paragraph{DeepCoder.} The DeepCoder task involves using DSL to perform operations on integer lists. In DeepCoder, each line represents a subroutine that performs atomic operations on previous variables and assigns the results to new variables. The result of the final line is the program's output. For example, program \texttt{a $\leftarrow$ [int] | b $\leftarrow$ FILTER(<0) a | c $\leftarrow$ MAP(*4) b | d $\leftarrow$ SORT c | e $\leftarrow$ REVERSE b} (where ``\texttt{|}'' denotes subroutine separator.) transforms the input \texttt{[-17, -3, 4, 11, 0, -5, -9, 13, 6, 6, -8, 11]} into the output \texttt{[-12, -20, -32, -36, -68]}. We provide detailed RobustFill and Deepcoder DSLs in Appendix~\ref{app:dsl}. 
\subsection{Deductive Code Reasoning}
Deductive code reasoning refers to the process of deriving a sound inference $\mathcal{O}$ by reasoning from the given premise $\mathcal{I}$, assuming the validity of the argument $\mathcal{P}$. Deductive code reasoning can be instantiated as an output prediction task~\citep{gu2024cruxeval}. Based on the given premise, the output prediction requires the LLM to simulate a compiler~\citep{kim2024llmcompiler}, executing step by step until it arrives at a valid conclusion.
For example, given a program \texttt{P = lambda text, value: ''.join(list(text) + [value])} and inputs \texttt{text = `bcksrut', b = `q'}, the output prediction from LLM should be \texttt{`bcksrutq'}.

\subsection{Abductive Code Reasoning}
Starting from existing facts $\mathcal{P}$ and $\mathcal{O}$, deriving the most reasonable and optimal explanation $\mathcal{I}$ is referred to as abductive code reasoning. This meta-benchmark can be framed as an input prediction task. Given the provided facts, the input prediction requires the LLM to backtrack through the program's execution process to recover the potential inputs. In cases where multiple possible inputs exist, the model should apply Occam's Razor and return the simplest input. For example, given a program \texttt{P = lambda nums: nums + [nums[i \% 2] for i in range(len(nums))]} and outputs \texttt{[-1, 0, 0, 1, 1, -1, 0, -1, 0, -1]}, the input prediction from LLM should be \texttt{[-1, 0, 0, 1, 1]}.

Deductive code and abductive code reasoning can be regarded as opposite processes; therefore, we selected two identical and representative datasets, CRUXEval~\citep{gu2024cruxeval} and LiveCodeBench~\citep{jain2024livecodebench}, as benchmarks to validate these two capabilities.

\paragraph{CRUXEval.} CRUXEval is a benchmark designed to evaluate code understanding and execution. Many models that achieve high scores on HumanEval~\citep{chen2021evaluating} do not show the same level of improvement on the CRUXEval benchmark. This benchmark includes 800 functions along with their corresponding inputs and outputs.

\paragraph{LiveCodeBench.} LiveCodeBench is a dynamically updated benchmark sourced from competition platforms. Each problem is timestamped, and we selected data from October 2023 (later than GPT-4o training) to March 2024 (the most recent), ensuring there is no data leakage and thereby guaranteeing the model's generalization performance.

\section{Code Reasoning with Hypothesis Decomposition and Amendment}
We aim to generate a reliable reasoning process for problem-solving by establishing a problem-solving pathway $f: \mathcal{X} \rightarrow \mathcal{Y}$. For a given task $\tau$ and the seen specifications/observations $\mathcal{X}^{s}_\tau$, the pathway $f$, should lead to a seen valid solution $\mathcal{Y}^{s}_\tau$ through a chain of reasoning.
We expect this pathway $f$ to have sufficient generalization capabilities to handle unseen specifications/observations $\mathcal{X}^{u}_\tau$.
To this aim, we employ a process involving hypothesis decomposition, execution verification, and amendment submission to iteratively explore and refine the reasoning pathway.
\begin{figure}[t]
    \centering
    \includegraphics[width=\linewidth]{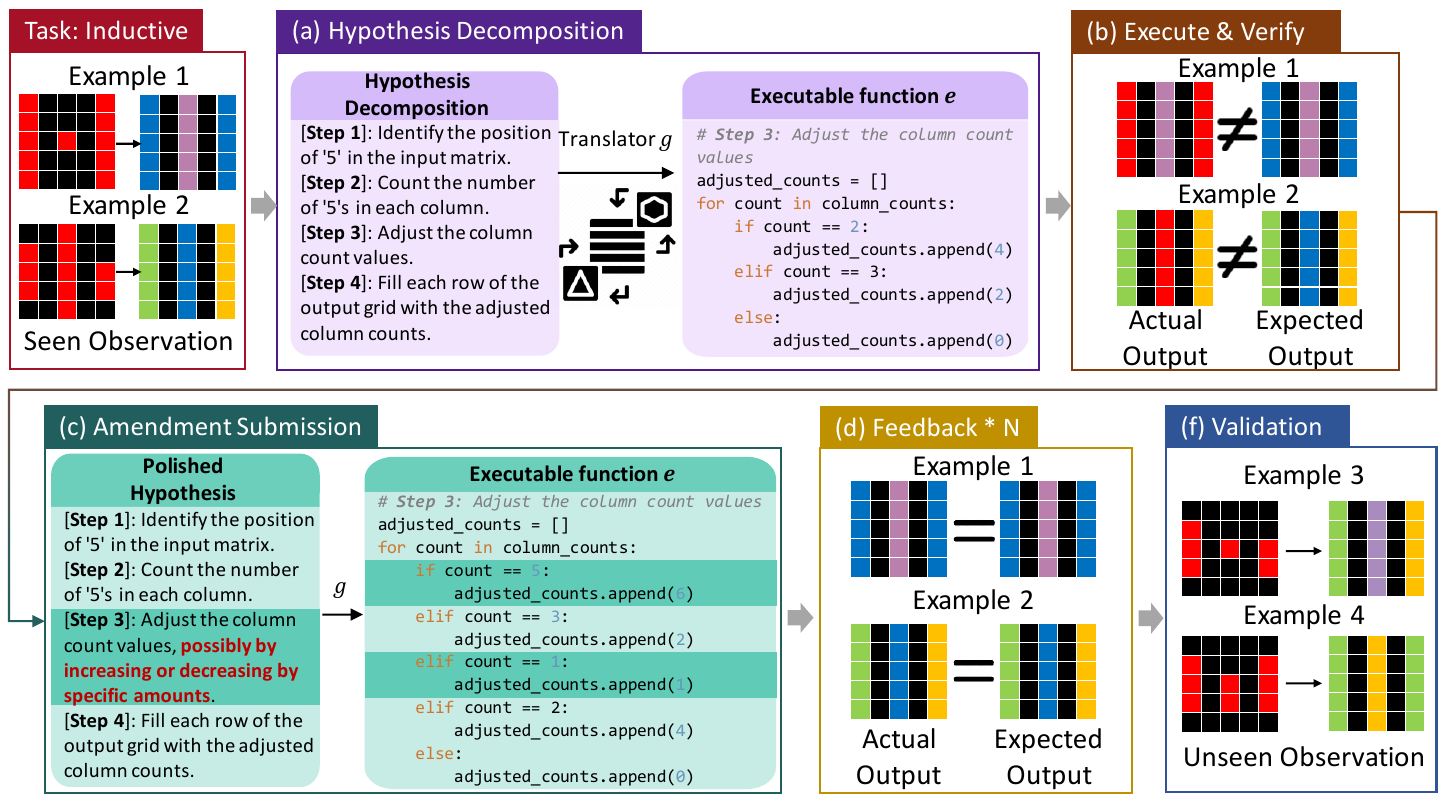}
    \caption{An overview of pipeline to solve code reasoning task. We decompose the hypothesis and generate executable functions step by step. After comparing the results with the seen observations and receiving feedback, we propose amendments, reflect on potential errors at each step, and generate revised hypotheses. This process is repeated until a valid problem-solving pathway is discovered. For concise expression, we show partial code snippets.}
    \label{fig:method}
    \vspace{-0.2cm}
\end{figure}
We first establish an initial hypothesis $h^0 \in \Sigma^*$ based on observations $x^{s}_\tau \in \mathcal{X}^{s}_\tau$, where $\Sigma^*$ is the closure form of LLM's vocabulary. This initial hypothesis $h^0$ serves as a preliminary solution pathway to the problem. Given the complexity of many problems, we decompose the hypothesis $h^0$ into simpler sub-hypotheses $h^0 \iff \{h^0_{s_0}, h^0_{s_1}, h^0_{s_2},...\}$. A translator function $g:\Sigma^* \rightarrow \Sigma_\mathcal{E}^*$, which maps the hypothesis space $\Sigma^*$ into an executable function space $\Sigma_\mathcal{E}^*$, is then used to `compiled' the sub-hypotheses $h^0$ into an executable function $e^0$. This executable function is directly applicable to the observations $x^{s}_\tau$, allowing for the derivation of conclusions $\widetilde{y}^{s}_\tau$, that is: 
\begin{equation}
    \widetilde{y}^{s}_\tau = g(h^0)({x}^{s}_\tau).
\end{equation}
Feedback $\mathcal{F}(y^{s}_\tau, \widetilde{y}^{s}_\tau)$ is used to evaluate the conclusions drawn from the current hypothesis, guiding the LLM to reflect on its sub-hypotheses. Through this iterative process of reflection, the model generates a new hypothesis $h^1$ for the next iteration. Finally, the problem-solving pathway $f$ is applied to unseen observations $\mathcal{X}^{u}_\tau$, and the model's generalization performance is assessed by measuring its accuracy:
\begin{equation}
    acc_\tau = \frac{1}{|\mathcal{X}^{u}_\tau|}\sum_{x^{u}_\tau \in \mathcal{X}^{u}_\tau}{\1\left[f(x^{u}_\tau)=y^{u}_\tau\right]}.
\end{equation}
The preceding section presents a unified framework for the hypothesis decomposition and amendment method. However, the implementation specifics differ across various tasks. In the following sections, we will introduce these task-specific variations in detail.

\paragraph{Hypothesis Decomposition.} We recognize that complex logical reasoning problems are difficult to encapsulate in a single reasonable hypothesis, which can adversely affect the performance of LLMs. Therefore, we require the LLM to decompose its hypotheses. Specifically, given an observation $x^{s}_\tau$, the LLM gradually presents corresponding hypotheses step by step.
For inductive code reasoning, $h_0$ represents the step-by-step hypothesis of the input-to-output transformation rules. For deductive and abductive code reasoning, $h_0$ refers to the step-by-step hypothesis regarding the functionality of the program.

\paragraph{Execution Verification.} After obtaining the hypothesis, we need to apply it to the observations. However, hypotheses are often not directly usable, so we need to convert the decomposed hypothesis into an executable function $e$ through a translator $g$. For inductive code reasoning, the executable function is a program; for deductive and abductive code reasoning, the executable function is the predicted output and input, respectively. These three types of task are then sent to a compiler to obtain the actual execution results, and the feedback generated by the compiler is provided to the LLM to help it further refine and adjust the sub-hypotheses.

\paragraph{Amendment Submission.} During the amendment submission stage, there are no significant differences in handling the three tasks. The LLM receives validation feedback from the tools and generates amendments based on this feedback, reflecting on possible issues in the previous hypotheses. The reflection process involves revising each sub-hypothesis individually, forming an updated hypothesis $h_1 \iff \{h^1_{s_0}, h^1_{s_1}, h^1_{s_2},...\}$. This process ensures that each sub-hypothesis is adjusted to better align with the observations and validation results, gradually improving the reasoning pathway's coherence and accuracy.

\section{Experiments}
\paragraph{Experimental Setup.} We utilize the latest and most advanced model, gpt-4o-2024-08-06, as the backbone LLM for all our experiments. We report the results using Llama-3.1-70B-Instruct, Qwen-max (qwen-max-2024-09-19)~\citep{bai2023qwen}, Claude 3.5 (claude-3-5-sonnet-20240620) in Appendix~\ref{app:more_llms}. Following the methodology of \citet{qiu2024phenomenal}, we set the temperature to 0.7. We report results using several methods: input-output (IO) prompting, standard prompting, Chain of Thought (CoT) \citep{wei2023chainofthought}, Program of Thought (PoT) \citep{chen2023programthought}, Chain of Code (CoC) \citep{li2024chaincode}, Self-Consistency (SC) \citep{wang2023selfconsistency} and Self-Refine (SR)~\citep{madaan2024self}, all implemented with 2-shot learning.\footnote{Not all methods are suitable for these three meta-benchmarks, thus we selected the most appropriate methods for each benchmark.} For our proposed process, we employ 0-shot prompts, allowing the LLM to explore problem-solving pathways in a more flexible manner. We provide detailed prompt templates in Appendix~\ref{app:prompts}.
\subsection{Inductive Code Reasoning}
\begin{table*}[t!]
\centering
\caption{RHDA method on inductive code reasoning task. $T$ refers to the maximum number of iterations. $N$ refers to the number of candidates.}
\scalebox{0.75}{
\begin{tabular}{lcccccccc}
\toprule
\multicolumn{1}{l}{\multirow{2}{*}{Method}} & \multicolumn{4}{c}{\textbf{Accuracy}} & \multicolumn{4}{c}{\textbf{Task Accuracy}} \\ \cmidrule(lr){2-5} \cmidrule(lr){6-9} 
\multicolumn{1}{c}{} & List Func & MiniARC & RobustFill & Deepcoder & List Func & MiniARC & RobustFill & Deepcoder \\ \midrule
IO & \textbf{64.85} & \textbf{28.21} & \textbf{61.74} & 23.78 & 38.00 & 13.08 & 21.74 & 10.42 \\
PoT & 44.90 & 10.90 & 37.39 & 30.90 & 33.60 & 8.46 & 21.74 & 19.79 \\
CoC & 42.45 & 10.90 & 31.30 & 26.39 & 34.40 & 4.62 & 13.04 & 13.54 \\
SC \scriptsize{(N=3)} & 52.95 & 12.31 & 46.09 & 37.85 & 41.20 & 9.23 & 26.09 & 26.04 \\ 
SR \scriptsize{(T=2)} & 51.10 & 10.26 & 41.74 & 36.81 & 41.60 & 8.46 & 21.74 & 25.00 \\ \midrule
w/o Sub-Hyp & 42.45 & 7.95 & 40.87 & 18.05 & 33.20 & 4.62 & 21.74 & 9.37 \\ w/o Amend & 47.10 & 8.46 & 35.65 & 30.21 & 36.40 & 6.92 & 17.39 & 19.79 \\ \midrule
T=2, N=1 & 51.05 & 12.56 & 43.48 & 38.89 & 41.20 & 10.77 & 30.43 & 23.96 \\
T=3, N=1 & 53.20 &  14.10 &  47.83 & 38.19  & 44.00 &  11.54 & 30.43  & 26.04  \\
T=2, N=3 & 58.35 & 19.74  & 54.78  & \textbf{43.06}  & \textbf{48.80} & \textbf{13.85}  &  \textbf{34.78} & \textbf{29.17}  \\
\bottomrule
\end{tabular}
}
\label{tab:in_main}
\end{table*}
For inductive code reasoning, we establish four baseline methods. The Input-Output (IO) prompting requires the LLM to predict outputs based on all seen observations and an unseen input. The Program of Thought (PoT) method generates and executes programs to derive outputs. The CoC method prompts the LLM to utilize pseudocode for reasoning in output prediction. The SC method builds upon PoT by sampling multiple programs and selecting the one that demonstrates optimal performance on seen observations.
Furthermore, since each example may contain multiple unseen observations, we adopt the approach from~\citep{qiu2024phenomenal} to define task accuracy externally. An example is deemed passed only when all unseen observations within it pass; thus, the proportion of passed examples reflects the task accuracy. The experimental results are presented in Table~\ref{tab:in_main}. 

The results demonstrate that the RHDA method achieves optimal performance across four benchmarks, with task accuracy exceeding that of the second-best methods by 18.45\%, 5.89\%, 33.31\%, and 12.02\%, respectively. However, we observe that RHDA appears to underperform compared to IO prompting. This is because the IO prompt does not generate a hypothesis that satisfies all observations but instead predicts the output for a single input. A successful prediction for a single instance does not generate a hypothesis that satisfies all observations, resulting in a high prediction accuracy but a relatively low task accuracy.

\paragraph{Ablation Study.}
We introduce two variants to separately validate the effectiveness of hypothesis decomposition and amendment submission. The first variant does not require the LLM to decompose hypotheses, referred to as w/o Sub-Hyp. The second variant, termed w/o Amend, indicates that the model no longer modifies its hypotheses through reflection.
The experimental results presented in Table~\ref{tab:in_main} show that the performance of these two variants declined by 25.39\% to 67.88\% and 19.28\% to 57.14\%, respectively. This finding suggests that the introduction of sub-hypotheses is a critical step, as it simplifies complex problems, reducing the workload for the subsequent translator $g$ while also enabling individual amendments to each sub-hypothesis. Nonetheless, the reflection process is equally important. Our results align with previous research~\citep{zhao2024repair, olausson2024repair, peng2023check} indicating that rational reflection can significantly enhance performance.
\subsection{Deductive Code Reasoning}
\begin{wraptable}{r}{0.47\textwidth}
\centering
\footnotesize
\vspace{-20pt}
\caption{RHDA method on deductive code reasoning task. $T$ refers to the maximum number of iterations. $N$ refers to the number of candidates.}
\begin{tabular}{lcc}
\toprule
  & CRUXEval & LiveCodeBench \\ \midrule
Standard    & 68.75  & 41.18  \\
CoT  & 89.12   & 83.14   \\
SC \scriptsize{(N=3)}   & 71.12    & 36.27   \\
SR \scriptsize{(T=2)}   & 80.38    & 63.73   \\
CoC  & 85.62    & 81.37  \\ \midrule
w/o Amend   & 86.62 & 71.29 \\
T=2, N=1 & \textbf{90.62}  & \textbf{84.16} \\ \bottomrule
\end{tabular}
\label{tab:de_main}
\end{wraptable}
For deductive code reasoning, we select standard prompting, CoT, SC, SR and CoC as benchmark methods. The experimental results are presented in Table~\ref{tab:de_main}. These results indicate that the CoT and CoC methods significantly enhanced the accuracy of reasoning outcomes by guiding the model to think step-by-step about function capabilities. Our proposed method advances this further, achieving optimal performance with a single round of amendments, resulting in an improvement of up to 104.37\% compared with baseline method. A horizontal comparison of the two datasets revealed that, due to the absence of LiveCodeBench data in internet corpora, the performance with standard prompts showed a marked advantage, with the SC method amplifying this gap. Notably, the combination of CoT, CoC, and hypothesis decomposition and amendment enabled the LLM to exhibit a substantial degree of reasoning and generalization ability, nearly solving all presented problems.

\subsection{Abductive Code Reasoning}
\begin{wrapfigure}{r}{0.5\textwidth}
    \vspace{-10pt}
    \centering
    \includegraphics[width=0.5\textwidth]{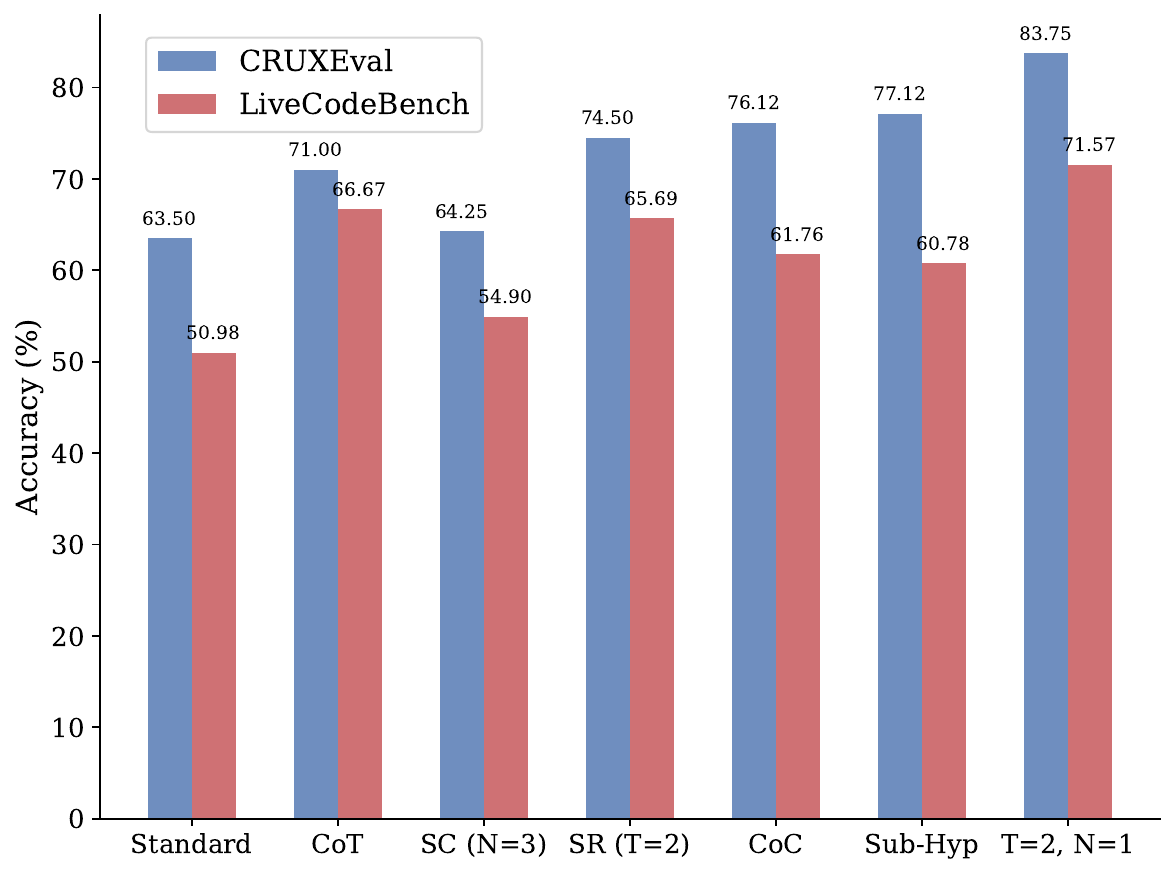}
    \caption{RHDA method on abductive code reasoning task. $T$ refers to the maximum number of iterations. $N$ refers to the number of candidates.}
    \vspace{-6pt}
    \label{fig:ab_result}
\end{wrapfigure}
For abductive code reasoning, we employ the same baseline methods as those used for deductive reasoning. The experimental results are presented in Figure~\ref{fig:ab_result}. Compared to deductive reasoning, abductive reasoning involves a reverse thinking process, which presents significant challenges. The LLM cannot derive the program's intermediate states through deduction and must first establish an abstract-level understanding of the function's behavior before proceeding with abduction.
On the CRUXEval dataset, the performance decline for abductive reasoning ranged from 8.20\% to 25.52\%. However, the hypothesis decomposition and amendment approach demonstrate robustness, as the change in reasoning modes resulted in only minimal performance degradation (8.20\%) while still outperforming baseline methods by 10.02\% to 31.89\% on the CRUXEval dataset and 7.35\% to 40.39\% on the LiveCodeBench dataset. A horizontal comparison of the two datasets revealed a trend similar to that observed in deductive reasoning, with an overall performance decline on the LiveCodeBench dataset, suggesting a complex relationship between reasoning and recall.
\subsection{Qualitative Analyze}
We select some cases to conduct an in-depth exploration of the quality of RHDA.
\begin{table*}[ht!]
    \caption{We compare the results obtained using the sub-hypothesis decomposition method with those obtained without it. The results without hypothesis decomposition are presented at the top of the table, while those with hypothesis decomposition are shown below. Benchmark: MiniARC-ID26.}
    \centering
    \scalebox{0.8}{
    \begin{tabular}{ccc}
\toprule
\textbf{Observations} & \textbf{Hypothesis} & \textbf{Executable Function} \\ \midrule
\multirow{2}{*}{\thead{\includegraphics[width=3.5cm]{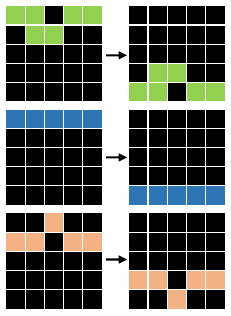}}} & \thead{\textbf{No Sub Hypothesis:}\\For a given 5x5 matrix input,\\ shift the first row to the last row,\\ the second row to the fourth row,\\ the third row to the third row \\(unchanged), the fourth row to the \\second row, and the fifth \\row to the first row. This rotates \\the rows up by two positions.
} & \thead{
\begin{lstlisting}[language=Python, style=mystyle]
def fn(matrix):
  if len(matrix) != 5 or any(len(row) != 5 for row in matrix):
    raise ValueError("Input must be a 5x5 matrix")
  # Create a new matrix to store the result
  result = [[0]*5 for _ in range(5)]
  # Apply the row shifting rule
  result[0] = matrix[2]  # Third row to first row
  result[1] = matrix[3]  # Fourth row to second row
  result[2] = matrix[4]  # Fifth row to third row
  result[3] = matrix[0]  # First row to fourth row
  result[4] = matrix[1]  # Second row to fifth row
  return result
\end{lstlisting}} \\ \cline{2-3} 
& \thead{\textbf{Sub Hypothesis:}\\Step 1: Identify the non-zero rows.\\Step 2: Move non-zero rows to the bottom.\\ Step 3: Shift all rows down to fill the grid.\\Step 4: Repeat steps for next input.} & \thead{
\begin{lstlisting}[language=Python, style=mystyle]
def fn(grid):
  # Step 1: Identify non-zero rows
  non_zero_rows = [row for row in grid if any(row)]
  # Step 2: Reverse the order of non-zero rows
  non_zero_rows.reverse()
  # Step 3: Calculate the number of zero rows needed
  num_zero_rows = len(grid) - len(non_zero_rows)
  # Step 4: Create the new grid with zero rows at the top
  zero_rows = [[0] * len(grid[0]) for _ in range(num_zero_rows)]
  # Step 5: Combine zero rows and reversed non-zero rows
  new_grid = zero_rows + non_zero_rows
  return new_grid
\end{lstlisting}} \\
\bottomrule
\end{tabular}
    }
    \label{tab:case_study_hyp1}
\vspace{-4mm}
\end{table*}


\paragraph{Hypothesis Decomposition Reduces Task Complexity.} In Table~\ref{tab:case_study_hyp1}, we present a qualitative analysis of the MiniARC benchmark, comparing the effects of using hypothesis decomposition versus not using it. As illustrated in the examples above, without hypothesis decomposition, the descriptions require a substantial amount of language, which ultimately leads to severe hallucinations in the program and results in failure. In contrast, the examples below, which utilize hypothesis decomposition, present clear and executable objectives, enabling the translator to easily write executable functions and achieve successful solutions as instructed.
\begin{table*}[ht!]
    \caption{We present the additional improvements resulting from the use of amendments. The results shown above are those obtained after the initial hypotheses, while the results displayed below reflect the outcomes following the submission of amendments and subsequent reflections. Benchmark: List Function-ID29.}
    \centering
    \scalebox{0.8}{
    \begin{tabular}{ccc}
\toprule
\textbf{Observations} & \textbf{Rounds} & \textbf{Executable Function} \\ \midrule
\multirow{2}{*}{\thead{\scriptsize{\texttt{[0, 8, 3, 9]}} $\rightarrow$ \scriptsize{{\texttt{[0, 8]}}} \\ \scriptsize{\texttt{[6, 1]}} $\rightarrow$ \scriptsize{\texttt{[]}}  \\ \scriptsize{\texttt{[4, 8, 7]}} $\rightarrow$ \scriptsize{\texttt{[4]}} \\ \dots }} &
\thead{\textbf{Round 1:} \\Step 1: Remove elements from the end \\ of the input list until a '9' is encountered. \\ Step 2: If '9' is present, remove all \\ elements after the last occurrence of '9'. \\ Step 3: If '9' is not present, remove the \\ last two elements of the list. \\ Step 4: If the list has fewer than two \\ elements, return an empty list.} & \thead{
\begin{lstlisting}[language=Python, style=mystyle]
def fn(lst):
  # Check if the list has fewer than two elements
  if len(lst)<2:
    return []
  # Check if '9' is present in the list
  if 9 in lst:
    # Find the last occurrence of '9'
    last_nine_index=len(lst)-1-lst[::-1].index(9)
    # Return the list up to and including the last '9'
    return lst[:last_nine_index+1]
  else:
    # If '9' is not present,
    # remove the last two elements
    return lst[:-2]
\end{lstlisting}} \\ \cline{2-3} & \thead{\textbf{Round 2:} \\Step 1: Remove the last two elements\\ from the input list. \\ Step 2: If the resulting list has fewer than \\ two elements, return an empty list.} & \thead{
\begin{lstlisting}[language=Python, style=mystyle]
def fn(input_list):
  # Remove the last two elements from the input list
  modified_list = input_list[:-2]
  # Check if the list has fewer than two elements
  if len(modified_list) < 2:
    return []
  return modified_list
\end{lstlisting}} \\
 \bottomrule
\end{tabular}
    }
    \label{tab:case_study_feed1)}
\vspace{-4mm}
\end{table*}

\paragraph{Amendments Guide LLM Towards Correct Pathway.} We present a qualitative analysis of the use of amendments in the List Function benchmark in Table~\ref{tab:case_study_feed1)}. The upper section displays the initialization of the hypothesis, where the LLM generates a potential guess based on the observations and translates it into an executable program. After offloading the execution to the tool (e.g., Python executor) and receiving feedback, amendments are proposed to modify the initial hypothesis. Following this reflection, the LLM re-optimizes the rules, ultimately yielding the correct execution results. More qualitative analyse examples please refer to Appendix~\ref{app:examples}.

\paragraph{Failure Analyse.} We also conduct an in-depth analysis of the reasons behind process failures in RHDA, detailed in Appendix~\ref{app:failure}. Our findings reveal that the primary limitation arises from the restricted intrinsic reasoning capabilities of LLMs, which continue to face challenges in understanding and addressing complex problems. These limitations are primarily reflected in two aspects:
\begin{itemize}
    \item Difficulty in Generating Accurate Sub-Hypotheses: The generation of sub-hypotheses during the reasoning process often proves inaccurate, leading to subsequent breakdowns in reasoning chains.
    \item Sensitivity to Initial Hypotheses: The model exhibits a pronounced dependency on its initial hypotheses. Even when feedback is provided through amendment submissions, the model struggles to break free from its original thought framework, constraining its reasoning capabilities.
\end{itemize}

\subsection{RHDA is a Flexible and Scalable Problem-solving Pathway}
\begin{figure}
    \centering
    \includegraphics[width=\textwidth]{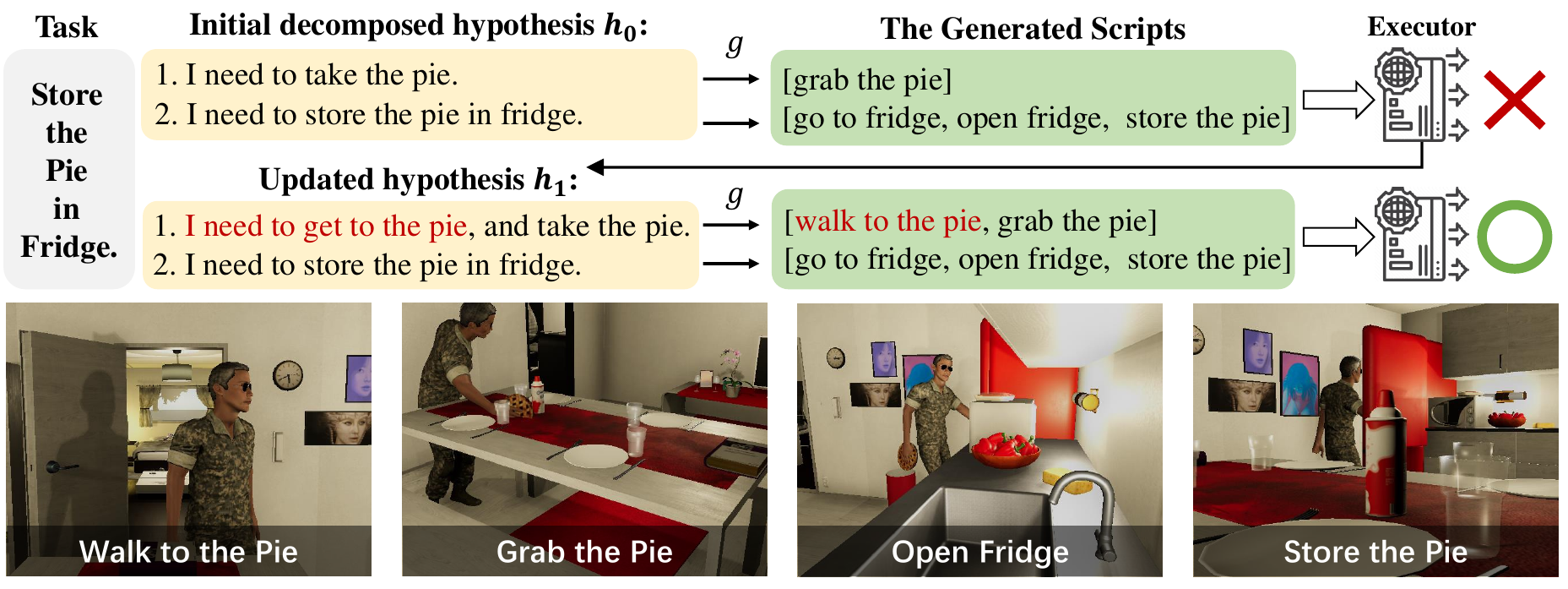}
    \caption{We demonstrate how RHDA can be extended to the VirtualHome framework to successfully complete the task of storing the pie in fridge.}
    \label{fig:virtualhome1}
\end{figure}

We consider extending the RHDA pipeline to more complex scenarios. To this end, we select VirtualHome~\citep{puig2018virtualhome, puig2020watchandhelp}, a sophisticated multi-agent platform for simulating household activities, as our new exploration subject. VirtualHome comprises a set of predefined atomic actions and objects that can be combined into high-level instructions. For example, `〈char0〉 [walk] 〈salmon〉' describes character 0 walking to the salmon. Given a specific scenario, the LLM is tasked with completing concrete housework using a series of high-level instructions. As depicted in Figure~\ref{fig:virtualhome1}, and guided by the RHDA process, we demonstrate how the LLM successfully accomplishes the task of storing pie in the fridge through the methods of hypothesis decomposition, execution verification (offloading to VirtualHome engine), and reflection. we show another example in Appneidx~\ref{app:virtualhome}.

\section{Limitation and Discussions}
\paragraph{Benchmark Selection.} This paper represents the first systematic exploration of the code reasoning task, focusing on the analysis of three forms of logical reasoning: inductive, deductive, and abductive. Due to time and cognitive constraints, we were unable to collect all benchmarks for testing. Our aim is to stimulate in-depth discussion on this topic and inspire meaningful follow-up research. While several excellent studies utilize code to address logical reasoning tasks~\citep{zelikman2023parsel, hu2023code, srivastava2024functional, liu2024codemind}, we did not include them here due to their differing starting points from this paper.
\paragraph{Hyperparameters.} The goal of this paper is to explore the potential of LLMs in code reasoning, rather than solely improving the performance of a specific code reasoning task. The RHDA framework serves as a preliminary exploration process; therefore, we didn't fully optimized the prompt templates or specific hyperparameters (such as temperature, $T$, and $N$) utilized. In the inductive code reasoning task, we examined a broader range of hyperparameter settings to illustrate that exploring multiple pathways aids in more effectively solving problems.
\paragraph{Task Assessment.} We propose a novel code reasoning task, and experimental results indicate that current state-of-the-art LLMs exhibit limitations in tackling this task. In the future, we aim to further explore this challenging area and investigate the boundaries of human capabilities in similar tasks.
\section{Related Work}
\paragraph{Reasoning with LLMs.}
LLMs such as GPT~\citep{openai2023gpt4}, LLaMA~\citep{touvron2023llama}, and Claude~\citep{anthropic2024claude}, demonstrate impressive reasoning capabilities across various NLP tasks~\citep{zhang2024llm_reasong_survey}. However, due to the problems of direct reasoning with LLMs such as hallucinations~\citep{ji2023survey_hallucination}, researchers have proposed several methods to enhance the reasoning power of LLMs. For example, 
\citep{zhouleast2most, xue2025decompose} decompose complex tasks into sequential subproblems, while 
\citep{sun2024adaplanner_feedback} refine reasoning through environment feedback. Moreover, intermediate representations, such as graphs~\citep{jiang2024resprompt_graph}, planning domain definition languages (PDDL)~\citep{guan2023leveraging_PDDL}, and triples~\citep{wang2023boosting_CoK}, have been employed to enhance LLM's reasoning.
Most recently, OpenAI o1~\citep{openai2024o1} demonstrates strong reasoning capabilities and broad world knowledge. Upon further contemplation, it is capable of reasoning through complex tasks and addressing challenges that exceed those faced by previous scientific, coding, and mathematical models.

Simultaneously, domain-specific reasoning with LLMs has gained attention. \citep{kim2024language_reason_computer} enhance reasoning outputs in computer tasks through recursive critique. In a case study using Minecraft, \citep{wang2023describe_reason_mc} introduce a Describe, Interpret, Plan, and Select framework for open-world multitasking. In computer vision, \citep{gupta2023visual_reason_cv} employ Python-like modular programs to tackle complex tasks. Nonetheless, reasoning in code remains an area yet to be thoroughly explored.

\paragraph{Improvement with Reflection.} Reflective ability is regarded as a crucial metric for evaluating LLMs as agents. Reflection can be categorized into internal and external based on its feedback source~\citep{pan2024automatically}. Internal reflection relies feedback from the model's own knowledge and parameters~\citep{huang2022large}, while external feedback comes from various sources, including humans~\citep{wang2023shepherd}, other models~\citep{paul2024refiner}, external tools~\citep{gou2024critic, chen2024teaching}, or knowledge bases~\citep{yao2023react, asai2024selfrag}.
\citep{huang2024large} find that LLMs struggle to self-correct their responses without external feedback, and in some cases, their performance may even decline following self-correction. Our work focuses on leveraging external tools, such as compilers, to generate feedback and enhance the performance of LLMs.

\section{Conclusion}
In this paper, we emphasized that the reasoning capabilities of LLMs still depend on recalling prior knowledge and highlighted that code reasoning has not been sufficiently explored as a novel perspective for examining the boundaries of LLM capabilities. Based on this consideration, we designed three meta-benchmarks—inductive code reasoning, deductive code reasoning, and abductive code reasoning—drawing on established forms of logical reasoning, and instantiated these benchmarks into eight specific tasks. Experimental results indicated that these benchmarks present significant challenges for current state-of-the-art LLMs.
To initially explore code reasoning tasks, we proposed a method involving \textbf{R}eflective \textbf{H}ypothesis \textbf{D}ecomposition and \textbf{A}mendment (\textbf{RHDA}). This method was iterative: LLMs need to generate decomposed initial hypotheses based on observations and employ a translator to interpret these into executable functions that can be directly applied to the observations. After obtaining the executable functions, we performed execution verification and submit amendments, allowing for reflection and refinement of the sub-hypotheses. Experimental results demonstrated that this approach, which integrated the principles of divide-and-conquer and reflection, can flexibly solve complex code reasoning problems, achieving performance improvements of 2 to 3 times compared to baseline methods. Finally, we extended this process to simulate household tasks in real-world complex scenarios to validate its scalability and transferability.
\section{Acknowledgment}
This research was partially supported by the Key Technologies R\&D Program of Anhui Province (No.202423k09020039), the National Natural Science Foundation of China (Grants No.62477044, 62406303), Anhui Provincial Natural Science Foundation (No. 2308085QF229), the Fundamental Research Funds for the Central Universities (No.WK2150110038, WK2150110034).
\section{Reproducibility Statement}
Our code, datasets and experimental results are available at \url{https://github.com/TnTWoW/code_reasoning}. Additionally, Appendix~\ref{app:prompts} contains details about pipeline and prompts used in method.

\bibliography{iclr2025_conference}

\begin{thebibliography}{76}
\providecommand{\natexlab}[1]{#1}
\providecommand{\url}[1]{\texttt{#1}}
\expandafter\ifx\csname urlstyle\endcsname\relax
  \providecommand{\doi}[1]{doi: #1}\else
  \providecommand{\doi}{doi: \begingroup \urlstyle{rm}\Url}\fi

\bibitem[Anthropic(2024)]{anthropic2024claude}
Anthropic.
\newblock Claude 3.5 sonnet, 2024.
\newblock URL \url{https://www.anthropic.com/news/claude-3-5-sonnet}.

\bibitem[Asai et~al.(2024)Asai, Wu, Wang, Sil, and Hajishirzi]{asai2024selfrag}
Akari Asai, Zeqiu Wu, Yizhong Wang, Avirup Sil, and Hannaneh Hajishirzi.
\newblock Self-{RAG}: Learning to retrieve, generate, and critique through self-reflection.
\newblock In \emph{The Twelfth International Conference on Learning Representations}, 2024.
\newblock URL \url{https://openreview.net/forum?id=hSyW5go0v8}.

\bibitem[Bai et~al.(2023)Bai, Bai, Chu, Cui, Dang, Deng, Fan, Ge, Han, Huang, et~al.]{bai2023qwen}
Jinze Bai, Shuai Bai, Yunfei Chu, Zeyu Cui, Kai Dang, Xiaodong Deng, Yang Fan, Wenbin Ge, Yu~Han, Fei Huang, et~al.
\newblock Qwen technical report.
\newblock \emph{arXiv preprint arXiv:2309.16609}, 2023.

\bibitem[Balog et~al.(2016)Balog, Gaunt, Brockschmidt, Nowozin, and Tarlow]{balog2016deepcoder}
Matej Balog, Alexander~L Gaunt, Marc Brockschmidt, Sebastian Nowozin, and Daniel Tarlow.
\newblock Deepcoder: Learning to write programs.
\newblock \emph{arXiv preprint arXiv:1611.01989}, 2016.

\bibitem[Bengio(2019)]{Bengio2019from}
Yoshua Bengio.
\newblock From system 1 deep learning to system 2 deep learning., 2019.
\newblock URL \url{https://nips.cc/Conferences/2019/Schedule?showEvent=15488}.

\bibitem[Brown et~al.(2020)Brown, Mann, Ryder, Subbiah, Kaplan, Dhariwal, Neelakantan, Shyam, Sastry, Askell, et~al.]{brown2020language}
Tom Brown, Benjamin Mann, Nick Ryder, Melanie Subbiah, Jared~D Kaplan, Prafulla Dhariwal, Arvind Neelakantan, Pranav Shyam, Girish Sastry, Amanda Askell, et~al.
\newblock Language models are few-shot learners.
\newblock \emph{Advances in neural information processing systems}, 33:\penalty0 1877--1901, 2020.

\bibitem[Chen et~al.(2021)]{chen2021evaluating}
Mark Chen et~al.
\newblock Evaluating large language models trained on code, 2021.
\newblock URL \url{https://arxiv.org/abs/2107.03374}.

\bibitem[Chen et~al.(2023)Chen, Ma, Wang, and Cohen]{chen2023programthought}
Wenhu Chen, Xueguang Ma, Xinyi Wang, and William~W. Cohen.
\newblock Program of thoughts prompting: Disentangling computation from reasoning for numerical reasoning tasks, 2023.
\newblock URL \url{https://arxiv.org/abs/2211.12588}.

\bibitem[Chen et~al.(2024)Chen, Lin, Sch{\"a}rli, and Zhou]{chen2024teaching}
Xinyun Chen, Maxwell Lin, Nathanael Sch{\"a}rli, and Denny Zhou.
\newblock Teaching large language models to self-debug.
\newblock In \emph{The Twelfth International Conference on Learning Representations}, 2024.
\newblock URL \url{https://openreview.net/forum?id=KuPixIqPiq}.

\bibitem[Chollet(2019)]{chollet2019measure}
François Chollet.
\newblock On the measure of intelligence, 2019.
\newblock URL \url{https://arxiv.org/abs/1911.01547}.

\bibitem[Devlin et~al.(2017)Devlin, Uesato, Bhupatiraju, Singh, Mohamed, and Kohli]{devlin2017robustfill}
Jacob Devlin, Jonathan Uesato, Surya Bhupatiraju, Rishabh Singh, Abdel-rahman Mohamed, and Pushmeet Kohli.
\newblock Robustfill: Neural program learning under noisy i/o.
\newblock In \emph{International conference on machine learning}, pp.\  990--998. PMLR, 2017.

\bibitem[Feeney \& Thompson(2014)Feeney and Thompson]{feeney2014reasoning}
A.~Feeney and V.A. Thompson.
\newblock \emph{Reasoning as Memory}.
\newblock Current Issues in Thinking and Reasoning. Taylor \& Francis, 2014.
\newblock ISBN 9781317820130.
\newblock URL \url{https://books.google.com.hk/books?id=gEuvBAAAQBAJ}.

\bibitem[Gou et~al.(2024)Gou, Shao, Gong, Shen, Yang, Duan, and Chen]{gou2024critic}
Zhibin Gou, Zhihong Shao, Yeyun Gong, Yelong Shen, Yujiu Yang, Nan Duan, and Weizhu Chen.
\newblock Critic: Large language models can self-correct with tool-interactive critiquing, 2024.
\newblock URL \url{https://arxiv.org/abs/2305.11738}.

\bibitem[Gu et~al.(2024)Gu, Rozière, Leather, Solar-Lezama, Synnaeve, and Wang]{gu2024cruxeval}
Alex Gu, Baptiste Rozière, Hugh Leather, Armando Solar-Lezama, Gabriel Synnaeve, and Sida~I. Wang.
\newblock Cruxeval: A benchmark for code reasoning, understanding and execution, 2024.
\newblock URL \url{https://arxiv.org/abs/2401.03065}.

\bibitem[Guan et~al.(2023)Guan, Valmeekam, Sreedharan, and Kambhampati]{guan2023leveraging_PDDL}
Lin Guan, Karthik Valmeekam, Sarath Sreedharan, and Subbarao Kambhampati.
\newblock Leveraging pre-trained large language models to construct and utilize world models for model-based task planning.
\newblock \emph{Advances in Neural Information Processing Systems}, 36:\penalty0 79081--79094, 2023.

\bibitem[Gulwani(2016)]{gulwani2016programming}
Sumit Gulwani.
\newblock Programming by examples-and its applications in data wrangling.
\newblock In \emph{Dependable Software Systems Engineering}, pp.\  137--158. IOS Press, 2016.

\bibitem[Gupta \& Kembhavi(2023)Gupta and Kembhavi]{gupta2023visual_reason_cv}
Tanmay Gupta and Aniruddha Kembhavi.
\newblock Visual programming: Compositional visual reasoning without training.
\newblock In \emph{Proceedings of the IEEE/CVF Conference on Computer Vision and Pattern Recognition}, pp.\  14953--14962, 2023.

\bibitem[Haidt(2001)]{haidt2001emotional}
Jonathan Haidt.
\newblock The emotional dog and its rational tail: a social intuitionist approach to moral judgment.
\newblock \emph{Psychological review}, 108\penalty0 (4):\penalty0 814, 2001.

\bibitem[Hardman \& Cowan(2015)Hardman and Cowan]{Kyle2015Reasoning}
Kyle Hardman and Nelson Cowan.
\newblock Reasoning and memory: People make varied use of the information available in working memory.
\newblock \emph{Journal of experimental psychology. Learning, memory, and cognition}, 42, 11 2015.
\newblock \doi{10.1037/xlm0000197}.

\bibitem[Hayes et~al.(2014)Hayes, Heit, and Rotello]{hayes2014memory}
Brett~K Hayes, Evan Heit, and Caren~M Rotello.
\newblock Memory, reasoning, and categorization: Parallels and common mechanisms.
\newblock \emph{Frontiers in psychology}, 5:\penalty0 529, 2014.

\bibitem[Heit et~al.(2012)Heit, Rotello, and Hayes]{Heit2012Relations}
Evan Heit, Caren~M. Rotello, and Brett~K. Hayes.
\newblock Chapter two - relations between memory and reasoning.
\newblock In Brian~H. Ross (ed.), \emph{The Psychology of Learning and Motivation}, volume~57 of \emph{Psychology of Learning and Motivation}, pp.\  57--101. Academic Press, 2012.
\newblock \doi{https://doi.org/10.1016/B978-0-12-394293-7.00002-9}.
\newblock URL \url{https://www.sciencedirect.com/science/article/pii/B9780123942937000029}.

\bibitem[Hu et~al.(2023)Hu, Yang, Lin, and Zhang]{hu2023code}
Yi~Hu, Haotong Yang, Zhouchen Lin, and Muhan Zhang.
\newblock Code prompting: a neural symbolic method for complex reasoning in large language models.
\newblock \emph{arXiv preprint arXiv:2305.18507}, 2023.

\bibitem[Huang et~al.(2022)Huang, Gu, Hou, Wu, Wang, Yu, and Han]{huang2022large}
Jiaxin Huang, Shixiang~Shane Gu, Le~Hou, Yuexin Wu, Xuezhi Wang, Hongkun Yu, and Jiawei Han.
\newblock Large language models can self-improve, 2022.
\newblock URL \url{https://arxiv.org/abs/2210.11610}.

\bibitem[Huang et~al.(2024)Huang, Chen, Mishra, Zheng, Yu, Song, and Zhou]{huang2024large}
Jie Huang, Xinyun Chen, Swaroop Mishra, Huaixiu~Steven Zheng, Adams~Wei Yu, Xinying Song, and Denny Zhou.
\newblock Large language models cannot self-correct reasoning yet.
\newblock In \emph{The Twelfth International Conference on Learning Representations}, 2024.
\newblock URL \url{https://openreview.net/forum?id=IkmD3fKBPQ}.

\bibitem[Jain et~al.(2024)Jain, Han, Gu, Li, Yan, Zhang, Wang, Solar-Lezama, Sen, and Stoica]{jain2024livecodebench}
Naman Jain, King Han, Alex Gu, Wen-Ding Li, Fanjia Yan, Tianjun Zhang, Sida Wang, Armando Solar-Lezama, Koushik Sen, and Ion Stoica.
\newblock Livecodebench: Holistic and contamination free evaluation of large language models for code.
\newblock \emph{arXiv preprint}, 2024.

\bibitem[Ji et~al.(2023)Ji, Lee, Frieske, Yu, Su, Xu, Ishii, Bang, Madotto, and Fung]{ji2023survey_hallucination}
Ziwei Ji, Nayeon Lee, Rita Frieske, Tiezheng Yu, Dan Su, Yan Xu, Etsuko Ishii, Ye~Jin Bang, Andrea Madotto, and Pascale Fung.
\newblock Survey of hallucination in natural language generation.
\newblock \emph{ACM Computing Surveys}, 55\penalty0 (12):\penalty0 1--38, 2023.

\bibitem[Jiang et~al.(2024)Jiang, Shakeri, Chan, Sanjabi, Firooz, Xia, Akyildiz, Sun, Li, Wang, et~al.]{jiang2024resprompt_graph}
Song Jiang, Zahra Shakeri, Aaron Chan, Maziar Sanjabi, Hamed Firooz, Yinglong Xia, Bugra Akyildiz, Yizhou Sun, Jinchao Li, Qifan Wang, et~al.
\newblock Resprompt: Residual connection prompting advances multi-step reasoning in large language models.
\newblock In \emph{Proceedings of the 2024 Conference of the North American Chapter of the Association for Computational Linguistics: Human Language Technologies (Volume 1: Long Papers)}, pp.\  5784--5809, 2024.

\bibitem[Kahneman(2011)]{kahneman2011thinking}
D.~Kahneman.
\newblock \emph{Thinking, Fast and Slow}.
\newblock Penguin Books Limited, 2011.
\newblock ISBN 9780141918921.
\newblock URL \url{https://books.google.com.hk/books?id=oV1tXT3HigoC}.

\bibitem[Kambhampati(2024)]{kambhampati2024can}
Subbarao Kambhampati.
\newblock Can large language models reason and plan?
\newblock \emph{Annals of the New York Academy of Sciences}, 1534\penalty0 (1):\penalty0 15--18, 2024.

\bibitem[Kim et~al.(2024{\natexlab{a}})Kim, Baldi, and McAleer]{kim2024language_reason_computer}
Geunwoo Kim, Pierre Baldi, and Stephen McAleer.
\newblock Language models can solve computer tasks.
\newblock \emph{Advances in Neural Information Processing Systems}, 36, 2024{\natexlab{a}}.

\bibitem[Kim et~al.(2024{\natexlab{b}})Kim, Moon, Tabrizi, Lee, Mahoney, Keutzer, and Gholami]{kim2024llmcompiler}
Sehoon Kim, Suhong Moon, Ryan Tabrizi, Nicholas Lee, Michael~W. Mahoney, Kurt Keutzer, and Amir Gholami.
\newblock An llm compiler for parallel function calling, 2024{\natexlab{b}}.
\newblock URL \url{https://arxiv.org/abs/2312.04511}.

\bibitem[Kim et~al.(2022)Kim, Phunyaphibarn, Ahn, and Kim]{kim2022playgrounds}
Subin Kim, Prin Phunyaphibarn, Donghyun Ahn, and Sundong Kim.
\newblock Playgrounds for abstraction and reasoning.
\newblock In \emph{NeurIPS 2022 Workshop on Neuro Causal and Symbolic AI (nCSI)}, 2022.

\bibitem[Kojima et~al.(2022)Kojima, Gu, Reid, Matsuo, and Iwasawa]{kojima2022large}
Takeshi Kojima, Shixiang~Shane Gu, Machel Reid, Yutaka Matsuo, and Yusuke Iwasawa.
\newblock Large language models are zero-shot reasoners.
\newblock \emph{Advances in neural information processing systems}, 35:\penalty0 22199--22213, 2022.

\bibitem[Kyllonen \& Christal(1990)Kyllonen and Christal]{Kyllonen1990Reasoning}
Patrick~C. Kyllonen and Raymond~E. Christal.
\newblock Reasoning ability is (little more than) working-memory capacity?!
\newblock \emph{Intelligence}, 14\penalty0 (4):\penalty0 389--433, 1990.
\newblock ISSN 0160-2896.
\newblock \doi{https://doi.org/10.1016/S0160-2896(05)80012-1}.
\newblock URL \url{https://www.sciencedirect.com/science/article/pii/S0160289605800121}.

\bibitem[Li et~al.(2024)Li, Liang, Zeng, Chen, Hausman, Sadigh, Levine, Fei-Fei, Xia, and Ichter]{li2024chaincode}
Chengshu Li, Jacky Liang, Andy Zeng, Xinyun Chen, Karol Hausman, Dorsa Sadigh, Sergey Levine, Li~Fei-Fei, Fei Xia, and Brian Ichter.
\newblock Chain of code: Reasoning with a language model-augmented code emulator, 2024.
\newblock URL \url{https://arxiv.org/abs/2312.04474}.

\bibitem[Lin et~al.(2024)Lin, Huang, Zhao, Chen, Liu, Lian, Li, and Wang]{lin2024learning}
Xin Lin, Zhenya Huang, Hongke Zhao, Enhong Chen, Qi~Liu, Defu Lian, Xin Li, and Hao Wang.
\newblock Learning relation-enhanced hierarchical solver for math word problems.
\newblock \emph{IEEE Transactions on Neural Networks and Learning Systems}, 35\penalty0 (10):\penalty0 13830--13844, 2024.

\bibitem[Liu et~al.(2024{\natexlab{a}})Liu, Zhang, Ibrahimzada, and Jabbarvand]{liu2024codemind}
Changshu Liu, Shizhuo~Dylan Zhang, Ali~Reza Ibrahimzada, and Reyhaneh Jabbarvand.
\newblock Codemind: A framework to challenge large language models for code reasoning.
\newblock \emph{arXiv preprint arXiv:2402.09664}, 2024{\natexlab{a}}.

\bibitem[Liu et~al.(2023{\natexlab{a}})Liu, Huang, Ma, Liu, Chen, Su, and Liu]{liu2023guiding}
Jiayu Liu, Zhenya Huang, Zhiyuan Ma, Qi~Liu, Enhong Chen, Tianhuang Su, and Haifeng Liu.
\newblock Guiding mathematical reasoning via mastering commonsense formula knowledge.
\newblock In \emph{Proceedings of the 29th ACM SIGKDD Conference on Knowledge Discovery and Data Mining}, KDD '23, pp.\  1477–1488. Association for Computing Machinery, 2023{\natexlab{a}}.
\newblock ISBN 9798400701030.
\newblock \doi{10.1145/3580305.3599375}.
\newblock URL \url{https://doi.org/10.1145/3580305.3599375}.

\bibitem[Liu et~al.(2023{\natexlab{b}})Liu, Huang, Zhai, and Liu]{liu2023learning}
Jiayu Liu, Zhenya Huang, Chengxiang Zhai, and Qi~Liu.
\newblock Learning by applying: A general framework for mathematical reasoning via enhancing explicit knowledge learning.
\newblock In \emph{Proceedings of the AAAI Conference on Artificial Intelligence}, volume~37, pp.\  4497--4506, 2023{\natexlab{b}}.

\bibitem[Liu et~al.(2024{\natexlab{b}})Liu, Huang, Xiao, Sha, Wu, Liu, Wang, and Chen]{liu2024socraticlm}
Jiayu Liu, Zhenya Huang, Tong Xiao, Jing Sha, Jinze Wu, Qi~Liu, Shijin Wang, and Enhong Chen.
\newblock Socratic{LM}: Exploring socratic personalized teaching with large language models.
\newblock In \emph{The Thirty-eighth Annual Conference on Neural Information Processing Systems}, 2024{\natexlab{b}}.
\newblock URL \url{https://openreview.net/forum?id=qkoZgJhxsA}.

\bibitem[Madaan et~al.(2024)Madaan, Tandon, Gupta, Hallinan, Gao, Wiegreffe, Alon, Dziri, Prabhumoye, Yang, et~al.]{madaan2024self}
Aman Madaan, Niket Tandon, Prakhar Gupta, Skyler Hallinan, Luyu Gao, Sarah Wiegreffe, Uri Alon, Nouha Dziri, Shrimai Prabhumoye, Yiming Yang, et~al.
\newblock Self-refine: Iterative refinement with self-feedback.
\newblock \emph{Advances in Neural Information Processing Systems}, 36, 2024.

\bibitem[Moskvichev et~al.(2023)Moskvichev, Odouard, and Mitchell]{moskvichev2023concept}
Arseny Moskvichev, Victor~Vikram Odouard, and Melanie Mitchell.
\newblock The conceptarc benchmark: Evaluating understanding and generalization in the arc domain, 2023.
\newblock URL \url{https://arxiv.org/abs/2305.07141}.

\bibitem[Olausson et~al.(2024)Olausson, Inala, Wang, Gao, and Solar-Lezama]{olausson2024repair}
Theo~X. Olausson, Jeevana~Priya Inala, Chenglong Wang, Jianfeng Gao, and Armando Solar-Lezama.
\newblock Is self-repair a silver bullet for code generation?
\newblock In \emph{International Conference on Learning Representations (ICLR)}, 2024.

\bibitem[OpenAI(2023)]{openai2023gpt4}
OpenAI.
\newblock Gpt-4 technical report, 2023.

\bibitem[OpenAI(2024)]{openai2024o1}
OpenAI.
\newblock Introducing openai o1, 2024.
\newblock URL \url{https://openai.com/o1/}.

\bibitem[Pan et~al.(2024)Pan, Saxon, Xu, Nathani, Wang, and Wang]{pan2024automatically}
Liangming Pan, Michael Saxon, Wenda Xu, Deepak Nathani, Xinyi Wang, and William~Yang Wang.
\newblock Automatically correcting large language models: Surveying the landscape of diverse automated correction strategies.
\newblock \emph{Transactions of the Association for Computational Linguistics}, 12:\penalty0 484--506, 2024.
\newblock \doi{10.1162/tacl_a_00660}.
\newblock URL \url{https://aclanthology.org/2024.tacl-1.27}.

\bibitem[Paul et~al.(2024)Paul, Ismayilzada, Peyrard, Borges, Bosselut, West, and Faltings]{paul2024refiner}
Debjit Paul, Mete Ismayilzada, Maxime Peyrard, Beatriz Borges, Antoine Bosselut, Robert West, and Boi Faltings.
\newblock {REFINER}: Reasoning feedback on intermediate representations.
\newblock In Yvette Graham and Matthew Purver (eds.), \emph{Proceedings of the 18th Conference of the European Chapter of the Association for Computational Linguistics (Volume 1: Long Papers)}, pp.\  1100--1126, St. Julian{'}s, Malta, March 2024. Association for Computational Linguistics.
\newblock URL \url{https://aclanthology.org/2024.eacl-long.67}.

\bibitem[Peng et~al.(2023)Peng, Galley, He, Cheng, Xie, Hu, Huang, Liden, Yu, Chen, et~al.]{peng2023check}
Baolin Peng, Michel Galley, Pengcheng He, Hao Cheng, Yujia Xie, Yu~Hu, Qiuyuan Huang, Lars Liden, Zhou Yu, Weizhu Chen, et~al.
\newblock Check your facts and try again: Improving large language models with external knowledge and automated feedback.
\newblock \emph{arXiv preprint arXiv:2302.12813}, 2023.

\bibitem[Puig et~al.(2018)Puig, Ra, Boben, Li, Wang, Fidler, and Torralba]{puig2018virtualhome}
Xavier Puig, Kevin Ra, Marko Boben, Jiaman Li, Tingwu Wang, Sanja Fidler, and Antonio Torralba.
\newblock Virtualhome: Simulating household activities via programs.
\newblock In \emph{Proceedings of the IEEE Conference on Computer Vision and Pattern Recognition}, pp.\  8494--8502, 2018.

\bibitem[Puig et~al.(2020)Puig, Shu, Li, Wang, Tenenbaum, Fidler, and Torralba]{puig2020watchandhelp}
Xavier Puig, Tianmin Shu, Shuang Li, Zilin Wang, Joshua~B. Tenenbaum, Sanja Fidler, and Antonio Torralba.
\newblock Watch-and-help: A challenge for social perception and human-ai collaboration, 2020.

\bibitem[Qiu et~al.(2024)Qiu, Jiang, Lu, Sclar, Pyatkin, Bhagavatula, Wang, Kim, Choi, Dziri, and Ren]{qiu2024phenomenal}
Linlu Qiu, Liwei Jiang, Ximing Lu, Melanie Sclar, Valentina Pyatkin, Chandra Bhagavatula, Bailin Wang, Yoon Kim, Yejin Choi, Nouha Dziri, and Xiang Ren.
\newblock Phenomenal yet puzzling: Testing inductive reasoning capabilities of language models with hypothesis refinement, 2024.
\newblock URL \url{https://arxiv.org/abs/2310.08559}.

\bibitem[Razeghi et~al.(2022)Razeghi, Logan~IV, Gardner, and Singh]{razeghi2022impact}
Yasaman Razeghi, Robert~L Logan~IV, Matt Gardner, and Sameer Singh.
\newblock Impact of pretraining term frequencies on few-shot numerical reasoning.
\newblock In \emph{Findings of the Association for Computational Linguistics: EMNLP 2022}, pp.\  840--854, 2022.

\bibitem[Rule(2020)]{rule2020child}
Joshua~Stewart Rule.
\newblock \emph{The child as hacker: building more human-like models of learning}.
\newblock PhD thesis, Massachusetts Institute of Technology, 2020.

\bibitem[Schaeffer et~al.(2024)Schaeffer, Miranda, and Koyejo]{schaeffer2024emergent}
Rylan Schaeffer, Brando Miranda, and Sanmi Koyejo.
\newblock Are emergent abilities of large language models a mirage?
\newblock \emph{Advances in Neural Information Processing Systems}, 36, 2024.

\bibitem[Shi et~al.(2024)Shi, Hong, Deng, Yin, Zaheer, and Sutton]{shi2024exedec}
Kensen Shi, Joey Hong, Yinlin Deng, Pengcheng Yin, Manzil Zaheer, and Charles Sutton.
\newblock {ExeDec}: Execution decomposition for compositional generalization in neural program synthesis.
\newblock In \emph{The Twelfth International Conference on Learning Representations}, 2024.

\bibitem[Srivastava et~al.(2024)Srivastava, B, au2, Menon, Sukumar, T, Philipose, Prince, and Thomas]{srivastava2024functional}
Saurabh Srivastava, Annarose~M B, Anto P~V au2, Shashank Menon, Ajay Sukumar, Adwaith~Samod T, Alan Philipose, Stevin Prince, and Sooraj Thomas.
\newblock Functional benchmarks for robust evaluation of reasoning performance, and the reasoning gap, 2024.

\bibitem[Strachan et~al.(2024)Strachan, Albergo, Borghini, Pansardi, Scaliti, Gupta, Saxena, Rufo, Panzeri, Manzi, et~al.]{strachan2024testing}
James~WA Strachan, Dalila Albergo, Giulia Borghini, Oriana Pansardi, Eugenio Scaliti, Saurabh Gupta, Krati Saxena, Alessandro Rufo, Stefano Panzeri, Guido Manzi, et~al.
\newblock Testing theory of mind in large language models and humans.
\newblock \emph{Nature Human Behaviour}, pp.\  1--11, 2024.

\bibitem[Sun et~al.(2024)Sun, Zhuang, Kong, Dai, and Zhang]{sun2024adaplanner_feedback}
Haotian Sun, Yuchen Zhuang, Lingkai Kong, Bo~Dai, and Chao Zhang.
\newblock Adaplanner: Adaptive planning from feedback with language models.
\newblock \emph{Advances in Neural Information Processing Systems}, 36, 2024.

\bibitem[Süß et~al.(2002)Süß, Oberauer, Wittmann, Wilhelm, and Schulze]{SU2002Working}
Heinz-Martin Süß, Klaus Oberauer, Werner~W Wittmann, Oliver Wilhelm, and Ralf Schulze.
\newblock Working-memory capacity explains reasoning ability—and a little bit more.
\newblock \emph{Intelligence}, 30\penalty0 (3):\penalty0 261--288, 2002.
\newblock ISSN 0160-2896.
\newblock \doi{https://doi.org/10.1016/S0160-2896(01)00100-3}.
\newblock URL \url{https://www.sciencedirect.com/science/article/pii/S0160289601001003}.

\bibitem[Touvron et~al.(2023)Touvron, Lavril, Izacard, Martinet, Lachaux, Lacroix, Rozi{\`e}re, Goyal, Hambro, Azhar, et~al.]{touvron2023llama}
Hugo Touvron, Thibaut Lavril, Gautier Izacard, Xavier Martinet, Marie-Anne Lachaux, Timoth{\'e}e Lacroix, Baptiste Rozi{\`e}re, Naman Goyal, Eric Hambro, Faisal Azhar, et~al.
\newblock Llama: Open and efficient foundation language models.
\newblock \emph{arXiv preprint arXiv:2302.13971}, 2023.

\bibitem[Wang et~al.(2023{\natexlab{a}})Wang, Sun, Li, and Gao]{wang2023boosting_CoK}
Jianing Wang, Qiushi Sun, Xiang Li, and Ming Gao.
\newblock Boosting language models reasoning with chain-of-knowledge prompting.
\newblock \emph{arXiv preprint arXiv:2306.06427}, 2023{\natexlab{a}}.

\bibitem[Wang et~al.(2023{\natexlab{b}})Wang, Yu, Tan, O'Brien, Pasunuru, Dwivedi-Yu, Golovneva, Zettlemoyer, Fazel-Zarandi, and Celikyilmaz]{wang2023shepherd}
Tianlu Wang, Ping Yu, Xiaoqing~Ellen Tan, Sean O'Brien, Ramakanth Pasunuru, Jane Dwivedi-Yu, Olga Golovneva, Luke Zettlemoyer, Maryam Fazel-Zarandi, and Asli Celikyilmaz.
\newblock Shepherd: A critic for language model generation, 2023{\natexlab{b}}.

\bibitem[Wang et~al.(2023{\natexlab{c}})Wang, Wei, Schuurmans, Le, Chi, Narang, Chowdhery, and Zhou]{wang2023selfconsistency}
Xuezhi Wang, Jason Wei, Dale Schuurmans, Quoc Le, Ed~Chi, Sharan Narang, Aakanksha Chowdhery, and Denny Zhou.
\newblock Self-consistency improves chain of thought reasoning in language models, 2023{\natexlab{c}}.
\newblock URL \url{https://arxiv.org/abs/2203.11171}.

\bibitem[Wang et~al.(2023{\natexlab{d}})Wang, Cai, Chen, Liu, Ma, Liang, and CraftJarvis]{wang2023describe_reason_mc}
Zihao Wang, Shaofei Cai, Guanzhou Chen, Anji Liu, Xiaojian Ma, Yitao Liang, and Team CraftJarvis.
\newblock Describe, explain, plan and select: interactive planning with large language models enables open-world multi-task agents.
\newblock In \emph{Proceedings of the 37th International Conference on Neural Information Processing Systems}, pp.\  34153--34189, 2023{\natexlab{d}}.

\bibitem[Wei et~al.(2022)Wei, Wang, Schuurmans, Bosma, Xia, Chi, Le, Zhou, et~al.]{wei2022chain}
Jason Wei, Xuezhi Wang, Dale Schuurmans, Maarten Bosma, Fei Xia, Ed~Chi, Quoc~V Le, Denny Zhou, et~al.
\newblock Chain-of-thought prompting elicits reasoning in large language models.
\newblock \emph{Advances in Neural Information Processing Systems}, 35:\penalty0 24824--24837, 2022.

\bibitem[Wei et~al.(2023)Wei, Wang, Schuurmans, Bosma, Ichter, Xia, Chi, Le, and Zhou]{wei2023chainofthought}
Jason Wei, Xuezhi Wang, Dale Schuurmans, Maarten Bosma, Brian Ichter, Fei Xia, Ed~Chi, Quoc Le, and Denny Zhou.
\newblock Chain-of-thought prompting elicits reasoning in large language models, 2023.
\newblock URL \url{https://arxiv.org/abs/2201.11903}.

\bibitem[Weston \& Sukhbaatar(2023)Weston and Sukhbaatar]{weston20232attention}
Jason Weston and Sainbayar Sukhbaatar.
\newblock System 2 attention (is something you might need too), 2023.
\newblock URL \url{https://arxiv.org/abs/2311.11829}.

\bibitem[Xie et~al.(2021)Xie, Raghunathan, Liang, and Ma]{xie2021explanation}
Sang~Michael Xie, Aditi Raghunathan, Percy Liang, and Tengyu Ma.
\newblock An explanation of in-context learning as implicit bayesian inference.
\newblock \emph{arXiv preprint arXiv:2111.02080}, 2021.

\bibitem[Xue et~al.(2025)Xue, Huang, Liu, Lin, Ning, Jin, Li, and Liu]{xue2025decompose}
Shangzi Xue, Zhenya Huang, Jiayu Liu, Xin Lin, Yuting Ning, Binbin Jin, Xin Li, and Qi~Liu.
\newblock Decompose, analyze and rethink: Solving intricate problems with human-like reasoning cycle.
\newblock \emph{Advances in Neural Information Processing Systems}, 37:\penalty0 357--385, 2025.

\bibitem[Yao et~al.(2023{\natexlab{a}})Yao, Yu, Zhao, Shafran, Griffiths, Cao, and Narasimhan]{yao2023tree}
Shunyu Yao, Dian Yu, Jeffrey Zhao, Izhak Shafran, Thomas~L. Griffiths, Yuan Cao, and Karthik Narasimhan.
\newblock {Tree of Thoughts}: Deliberate problem solving with large language models, 2023{\natexlab{a}}.

\bibitem[Yao et~al.(2023{\natexlab{b}})Yao, Zhao, Yu, Du, Shafran, Narasimhan, and Cao]{yao2023react}
Shunyu Yao, Jeffrey Zhao, Dian Yu, Nan Du, Izhak Shafran, Karthik Narasimhan, and Yuan Cao.
\newblock {ReAct}: Synergizing reasoning and acting in language models.
\newblock In \emph{International Conference on Learning Representations (ICLR)}, 2023{\natexlab{b}}.

\bibitem[Zeithamova et~al.(2012)Zeithamova, Schlichting, and Preston]{zeithamova2012hippocampus}
Dagmar Zeithamova, Margaret~L Schlichting, and Alison~R Preston.
\newblock The hippocampus and inferential reasoning: building memories to navigate future decisions.
\newblock \emph{Frontiers in human neuroscience}, 6:\penalty0 70, 2012.

\bibitem[Zelikman et~al.(2023)Zelikman, Huang, Poesia, Goodman, and Haber]{zelikman2023parsel}
Eric Zelikman, Qian Huang, Gabriel Poesia, Noah Goodman, and Nick Haber.
\newblock Parsel: Algorithmic reasoning with language models by composing decompositions.
\newblock In \emph{Thirty-seventh Conference on Neural Information Processing Systems}, 2023.
\newblock URL \url{https://openreview.net/forum?id=qd9qcbVAwQ}.

\bibitem[Zhang et~al.(2024)Zhang, Mao, Ge, Wang, de~Wynter, Xia, Wu, Song, Lan, and Wei]{zhang2024llm_reasong_survey}
Yadong Zhang, Shaoguang Mao, Tao Ge, Xun Wang, Adrian de~Wynter, Yan Xia, Wenshan Wu, Ting Song, Man Lan, and Furu Wei.
\newblock Llm as a mastermind: A survey of strategic reasoning with large language models.
\newblock \emph{arXiv preprint arXiv:2404.01230}, 2024.

\bibitem[Zhao et~al.(2024)Zhao, Huang, Ma, Li, Zhang, Jiang, Liu, Zhu, and Su]{zhao2024repair}
Yuze Zhao, Zhenya Huang, Yixiao Ma, Rui Li, Kai Zhang, Hao Jiang, Qi~Liu, Linbo Zhu, and Yu~Su.
\newblock {R}e{P}air: Automated program repair with process-based feedback.
\newblock In Lun-Wei Ku, Andre Martins, and Vivek Srikumar (eds.), \emph{Findings of the Association for Computational Linguistics ACL 2024}, pp.\  16415--16429, Bangkok, Thailand and virtual meeting, August 2024. Association for Computational Linguistics.
\newblock \doi{10.18653/v1/2024.findings-acl.973}.
\newblock URL \url{https://aclanthology.org/2024.findings-acl.973}.

\bibitem[Zhou et~al.(2023)Zhou, Sch{\"a}rli, Hou, Wei, Scales, Wang, Schuurmans, Cui, Bousquet, Le, et~al.]{zhouleast2most}
Denny Zhou, Nathanael Sch{\"a}rli, Le~Hou, Jason Wei, Nathan Scales, Xuezhi Wang, Dale Schuurmans, Claire Cui, Olivier Bousquet, Quoc~V Le, et~al.
\newblock Least-to-most prompting enables complex reasoning in large language models.
\newblock In \emph{The Eleventh International Conference on Learning Representations}, 2023.

\end{thebibliography}
\bibliographystyle{iclr2025_conference}

\clearpage
\appendix

\section{DSL Grammars}
\label{app:dsl}
RobustFill is a string manipulation task using the DSL. Figure~\ref{fig:robustfill_dsl} illustrates the DSL syntax for RobustFill. Our implementation is based on the works of ExeDec~\citep{shi2024exedec} and RobustFill~\citep{devlin2017robustfill}.

Deepcoder is a list transformation task using the DSL. Figure~\ref{fig:deepcoder_dsl}. This implementation is based on the works of ExeDec~\citep{shi2024exedec} and DeepCoder~\citep{balog2016deepcoder}.
\begin{figure*}[ht]
\small
\begin{alignat*}{2}
\mbox{Program } P\quad &:= &\quad& \T{Concat}(e_1, e_2, \hdots) \\
\mbox{Expression } e\quad &:= && s \logicalOR m \logicalOR o \logicalOR \T{ConstStr}(c) \\
\mbox{Compose } o\quad &:= && m_1(m_2) \logicalOR m(s) \\
\mbox{Substring } s\quad &:= && \T{SubStr}(k_1, k_2) \logicalOR 
\T{GetSpan}(r_1, i_1, b_1, r_2, i_2, b_2) \logicalOR \T{GetToken}(r, i) \\
&&& \logicalOR \T{GetUpto}(r) 
\logicalOR \T{GetFrom}(r)  \\
\mbox{Modification } m\quad &:= && \T{ToCase}(a) \logicalOR \T{Replace}(\delta_1, \delta_2) \logicalOR \T{Trim}() \logicalOR \T{GetFirst}(r, i) \logicalOR \T{GetAll}(r) \\
&&&\logicalOR  \T{Substitute}(r, i, c) \logicalOR  \T{SubstituteAll}(r, c) \logicalOR \T{Remove}(r, i) \logicalOR \T{RemoveAll}(r) \\
\mbox{Regex } r\quad &:= &&  \T{NUMBER} \logicalOR \T{WORD} \logicalOR \T{ALPHANUM} \logicalOR \T{ALL\_CAPS} \logicalOR \T{PROPER\_CASE} \logicalOR \T{LOWER} \logicalOR \T{DIGIT} \logicalOR \T{CHAR} \logicalOR \delta \\
\mbox{Case } a\quad &:= && \T{ALL\_CAPS} \logicalOR \T{PROPER\_CASE} \logicalOR \T{LOWER} \\
\mbox{Position } k\quad &:= && -100 \logicalOR -99 \logicalOR \hdots \logicalOR -1 \logicalOR 0 \logicalOR 1 \logicalOR 2 \logicalOR \hdots \logicalOR 100 \\
\mbox{Index } i\quad &:= && -5 \logicalOR -4 \logicalOR \hdots \logicalOR -1 \logicalOR 1 \logicalOR 2 \logicalOR \hdots \logicalOR 5 \\
\mbox{Boundary } b\quad &:= && \T{START} \logicalOR \T{END} \\
\mbox{Character } c\quad &:= && A\logicalOR\dots\logicalOR Z \logicalOR a\logicalOR\dots\logicalOR z \logicalOR 0\logicalOR\dots\logicalOR 9 \logicalOR \delta \\
\mbox{Delimiter } \delta\quad &:= && \texttt{\&,.?!@()[]\%\string{\string}/:;\$\# "'}
\end{alignat*}
    \caption{The DSL syntax for string manipulation tasks in the RobustFill domain.} 
\label{fig:robustfill_dsl}
\end{figure*}

\begin{figure*}[ht]
\small
\begin{alignat*}{2}
\mbox{Program } P\quad &:= &\quad& i_1; i_2; \hdots; a_1; a_2; \hdots \\
\mbox{Initialization } i\quad &:= && v \leftarrow \T{INPUT} \\
\mbox{Assignment } a\quad &:= && v \leftarrow f \logicalOR v \leftarrow h \\
\mbox{First-Order Operation } f\quad &:= && \T{Head}(l) \logicalOR \T{Last}(l) \logicalOR \T{Access}(n, l) \logicalOR \T{Minimum}(l) \logicalOR \T{Maximum}(l) \logicalOR \T{Sum}(l) \\
&&& \logicalOR \T{Take}(n, l) \logicalOR \T{Drop}(n, l) \logicalOR \T{Reverse}(l) \logicalOR \T{Sort}(l) \\
\mbox{Higher-Order Operation } h\quad &:= && \T{Map}(\lambda, l) \logicalOR \T{Filter}(\beta, l) \logicalOR \T{Count}(\beta, l) \logicalOR \T{ZipWith}(\Sigma, l, l) \logicalOR \T{Scanl1}(\Sigma, l) \\
\mbox{int $\rightarrow$ int Lambda } \lambda\quad &:= && (+1) \logicalOR (-1) \logicalOR (*2) \logicalOR (/2) \logicalOR (*(-1)) \logicalOR (*\!*\!2) \logicalOR (*3) \logicalOR (/3) \logicalOR (*4) \logicalOR (/4) \\
\mbox{int $\rightarrow$ bool Lambda } \beta\quad &:= && (>0) \logicalOR (<0) \logicalOR (\%2==0) \logicalOR (\%2==1) \\
\mbox{(int, int) $\rightarrow$ int Lambda } \Sigma\quad &:= && (+) \logicalOR (-) \logicalOR(*) \logicalOR (\min) \logicalOR (\max) \\
\mbox{Integer Variable } n\quad &:= && v \\
\mbox{List Variable } l\quad &:= && v \\
\mbox{Variable Name } v\quad &:= && x_1  \logicalOR x_2 \logicalOR \hdots
\end{alignat*}
    \caption{The DSL for integer and list manipulation tasks in the DeepCoder domain.} 
\label{fig:deepcoder_dsl}
\end{figure*}

\section{Experimental Results Using More LLMs}
\label{app:more_llms}
\begin{table*}[ht!]
\centering
\caption{Performance comparison of Llama3.1-70B-Instruct, Qwen-max, Claude 3.5 and GPT-4o on the PoT and RHDA methods in inductive code reasoning task. $T$ refers to the maximum number of iterations. $N$ refers to the number of candidates.}
\scalebox{0.72}{
\begin{tabular}{clcccccccc}
\toprule
\multicolumn{1}{l}{}    &          & \multicolumn{4}{c}{Accuracy}  & \multicolumn{4}{c}{Task Accuracy} \\ \cmidrule(lr){3-6} \cmidrule(lr){7-10}
Model                       & Method & MiniARC & List Func & RobustFill & DeepCoder & MiniARC & List Func & RobustFill & DeepCoder \\ \midrule
\multirow{3}{*}{Llama3.1} & PoT    & 3.08    & 35.25     & 14.78      & 22.92     & 1.54    & 26.80     & 8.70       & 11.46     \\
                          & Sub-Hyp  & 3.33 & 26.45 & 13.04 & 18.06 & 3.08   & 20.40  & 4.35   & 6.25   \\
                          & T=2, N=1 & 3.85 & 32.35 & 20.87 & 11.46 & 3.85   & 26.40  & 13.04  & 7.29  \\ \midrule
\multirow{3}{*}{Qwen-max} & PoT      & 6.41 & 41.75 & 36.52 & 25.35 & 3.85   & 30.00  & 21.74  & 14.58  \\
                          & Sub-Hyp  & 5.90 & 46.25 & 26.09 & 17.36 & 3.08   & 36.40  & 8.70   & 5.21   \\
                          & T=2, N=1 & 6.41 & 46.60 & 33.91 & 24.64 & 3.08   & 41.60  & 13.04  & 10.42 \\ \midrule
\multirow{3}{*}{Claude-3.5} & PoT    & 11.79   & 51.30     & 30.43      & 25.69     & 8.46    & 39.20     & 27.14      & 13.54     \\
                        & Sub-Hyp  & 12.56 & 53.55 & 22.61 & 33.33 & 9.23   & 42.40  & 8.70   & 16.67  \\
                        & T=2, N=1 & \textbf{18.21} & \textbf{57.95} & 33.91 & 29.86 & \textbf{13.85}  & \textbf{48.40}  & 17.39  & 20.83  \\ \midrule
\multirow{3}{*}{GPT-4o} & PoT      & 10.90 & 44.90 & 37.39 & 30.90 & 8.46   & 33.60  & 26.09  & 19.79  \\
                        & Sub-Hyp  & 8.46  & 47.10 & 35.65 & 24.65 & 6.92   & 36.40  & 17.39  & 12.50  \\
                        & T=2, N=1 & 12.56 & 51.05 & \textbf{43.48} & \textbf{38.89} & 10.77  & 41.20  & \textbf{40.43}  & \textbf{23.96}  \\ \bottomrule
\end{tabular}
}
\label{tab:in_main_claude}
\end{table*}

\begin{table*}[ht!]
\centering
\caption{Performance comparison of Llama3.1-70B-Instruct, Qwen-max, Claude 3.5 and GPT-4o on the CoT and RHDA methods in deductive and abductive code reasoning tasks. $T$ refers to the maximum number of iterations. $N$ refers to the number of candidates.}
\scalebox{0.9}{
\begin{tabular}{clcccc}
\toprule
\multicolumn{1}{l}{}        &          & \multicolumn{2}{c}{Deductive} & \multicolumn{2}{c}{Abductive} \\ \cmidrule(lr){3-4} \cmidrule(lr){5-6}
Model                       & Method   & CRUXEval    & LiveCodeBench   & CRUXEval    & LiveCodeBench   \\ \midrule
\multirow{3}{*}{Llama3.1} & CoT      & 40.25       & 7.84            & 53.12       & 38.24           \\
                          & Sub-Hyp  & 30.75       & 6.86            & 50.88       & 8.82            \\
                          & T=2, N=1 & 45.62       & 10.78           & 59.62       & 40.20   \\ \midrule
\multirow{3}{*}{Qwen-max} & CoT      & 81.12       & 86.27           & 75.12       & 58.82           \\
                          & Sub-Hyp  & 78.25       & 81.37           & 72.25       & 59.80           \\
                          & T=2, N=1 & 81.62       & \textbf{88.24}           & 79.38       & 66.67   \\ \midrule       
\multirow{3}{*}{Claude-3.5} & CoT      & 82.75       & 77.45           & 73.62       & 61.76           \\
                            & Sub-Hyp  & 77.75       & 65.69           & 74.75       & 53.92           \\
                            & T=2, N=1 & 86.88       & 80.39           & 83.38       & 61.76           \\ \midrule
\multirow{3}{*}{GPT-4o}     & CoT      & 89.12       & 83.14           & 71.00       & 66.67           \\
                            & Sub-Hyp  & 86.62       & 71.29           & 77.12       & 60.78           \\
                            & T=2, N=1 & \textbf{90.62}       & 84.16           & \textbf{83.75}       & \textbf{71.57}           \\ \bottomrule
\end{tabular}
}
\label{tab:de_ab_main_claude}
\end{table*}
We report the performance of Llama3.1-70B-Instruct, Qwen-max (qwen-max-2024-09-19), Claude 3.5 (claude-3-5-sonnet-20240620) using the RHDA method and compare them with GPT-4o (gpt-4o-2024-0806). The results for inductive code reasoning are shown in Table~\ref{tab:in_main_claude}. The experimental results indicate that GPT-4o performs better in solving DSL problems, while Claude 3.5 excels in General Propose Language (GPL) tasks. Compared to closed-source models, the open-source model Llama still exhibits relatively limited reasoning capabilities. However, in list manipulation tasks (List Function and Deepcoder), Llama demonstrates stronger programming abilities.
In Table~\ref{tab:de_ab_main_claude}, we report the performance of the models in deductive and abductive code reasoning together. The experimental results show that GPT-4o outperforms Claude 3.5 in terms of program understanding and execution capabilities. These results suggest that RHDA is a framework-agnostic general process that can achieve optimal performance through a single reflection, applicable to both Llama, Qwen, Claude and GPT series models.

\section{Benchmark Details}
\label{app:benchmark_details}
\begin{wraptable}{r}{0.47\textwidth}
\vspace{-7.7mm}
\footnotesize
\centering
\caption{The number of tasks per dataset, the numbers of seen examples per task, and unseen examples per task.}
\begin{tabular}{lccc}
\toprule
Dataset       & \# Tasks & \# Seen & \# Unseen \\ \midrule
List Function & 250      & 8       & 8         \\
MiniARC       & 130      & 3       & 3         \\
RobustFill    & 22       & 5       & 5         \\
Deepcoder     & 96       & 3       & 3         \\
CRUXEval      & 800      & 1       & 1         \\
LiveCodeBench & 102      & 1       & 1         \\ \bottomrule
\end{tabular}
\vspace{-2mm}
\label{tab:datasets}
\end{wraptable}
\paragraph{List Function.} We use the original dataset (Rule, 2020), which consists of a total of 250 tasks.
Due to the limited context lengths of LMs, we only use the first 16 examples from BIG-Bench (bench authors, 2023): 8 for seen examples and 8 for unseen examples. We manually examined the exemplars and found 8 examples are generally sufficient to describe the pattern.

\paragraph{MiniARC.} We use the data from ~\citep{qiu2024phenomenal}. Such tasks are typically difficult to describe in natural language at an
abstract level. Therefore, we did not consider them for our evaluations. As we only evaluate textonly models, we use textual representations of the original visual grids by mapping each cell to a corresponding integer.

\paragraph{RobustFill.} RobustFill is a string manipulation task where the model is expected to perform a combination of atomic operations, such as extracting a substring from position $k_1$ to $k_2$ using $SubString(k_1, k_1)$, to achieve generalization.
As an example, a program \texttt{ToCase(Lower, SubStr(1,3))}  converts full month names (January, April) to their abbreviations (jan, apr).

\paragraph{DeepCoder.} The DeepCoder task involves using DSL to perform operations on integer lists. In DeepCoder, each line represents a subroutine that performs atomic operations on previous variables and assigns the results to new variables. The result of the final line is the program's output. For example, program \texttt{a $\leftarrow$ [int] | b $\leftarrow$ FILTER(<0) a | c $\leftarrow$ MAP(*4) b | d $\leftarrow$ SORT c | e $\leftarrow$ REVERSE b} (where ``\texttt{|}'' denotes subroutine separator.) transforms the input \texttt{[-17, -3, 4, 11, 0, -5, -9, 13, 6, 6, -8, 11]} into the output \texttt{[-12, -20, -32, -36, -68]}

\section{RHDA Acting as an Agent in VirtualHome}
\label{app:virtualhome}
\begin{figure}[ht!]
    \centering
    \includegraphics[width=0.97\textwidth]{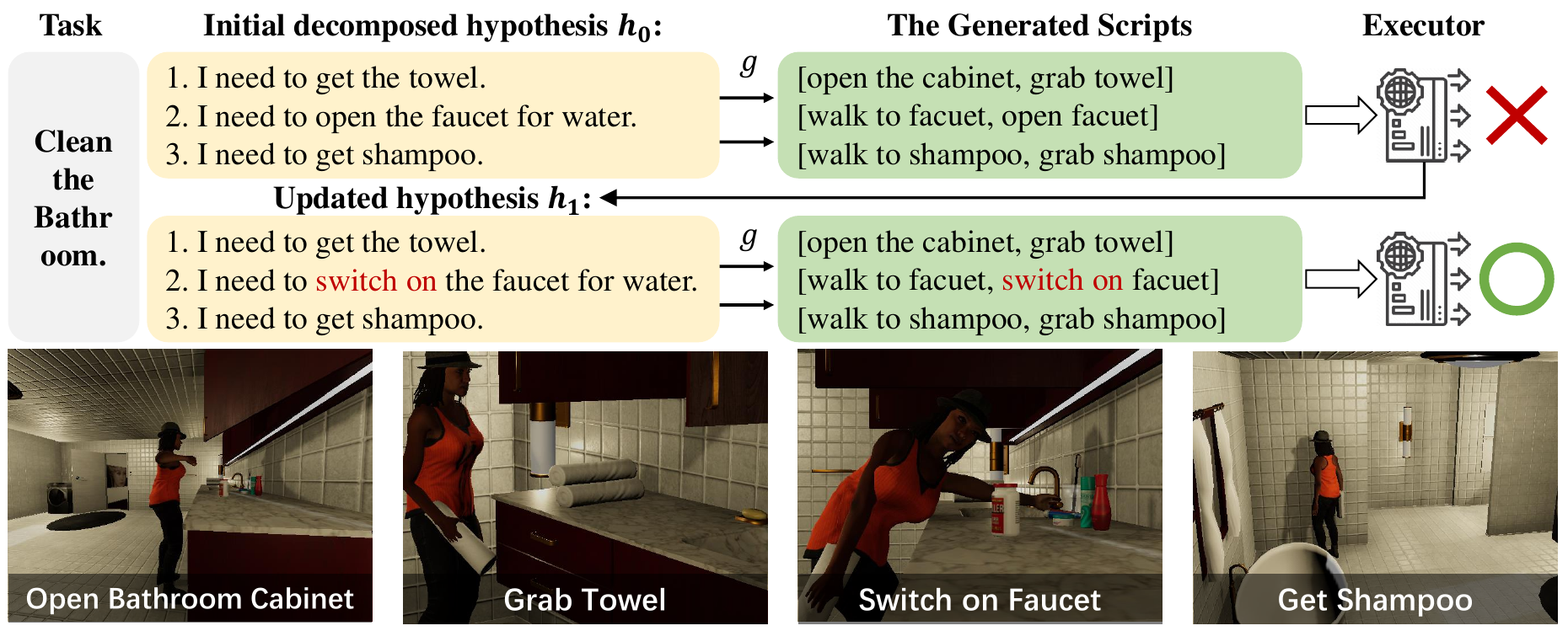}
    \caption{We illustrate how the RHDA framework can be extended to the VirtualHome environment to effectively accomplish the task of cleaning the bathroom.}
    \label{fig:virtualhome2}
\end{figure}
We utilized the RHDA framework to drive agent actions in the VirtualHome environment powered by LLMs. Figure~\ref{fig:virtualhome2} illustrates a task of cleaning the bathroom.

\begin{table}[ht!]
\caption{Execution Error Rate on VirtualHome}
\centering
\begin{tabular}{lcccc}
\toprule
 & \multicolumn{1}{l}{native GPT-4o} & w/o Sub-Hyp & \multicolumn{1}{l}{w/o Amend} & \multicolumn{1}{l}{RHDA} \\ \midrule
\# Error Action $\downarrow$  & 92   & 84   & 84   & \textbf{52}   \\
Avg. Err per Step $\downarrow$& 0.84 & 0.35 & 0.20 & \textbf{0.16} \\
Avg. Err per Task $\downarrow$ & 2.09 & 1.83 & 1.75 & \textbf{1.08} \\ \bottomrule
\end{tabular}
\label{tab:VH_quan}
\end{table}

We also provided some quantitative metrics to validate the potential of RHDA as a agent in VirtualHome. Specifically, we selected a total of 52 tasks across two scenarios in VirtualHome and manually tested their execution error rates. The test results are shown in Table~\ref{tab:VH_quan}, which indicate that native GPT-4o struggles to handle simulated real-world scenarios effectively. The primary cause of failure lies in generating scripts that, while semantically similar to correct actions, are not executable within the environment (e.g., `open the tap' is invalid action, whereas `touch the tap' is valid action). By employing the RHDA method, which incorporates step-by-step solutions and effective feedback mechanisms, the error rate was significantly reduced.

\section{Examples Analyses}
\subsection{Effective Case Study}
\label{app:examples}

\begin{table*}[ht!]
    \caption{We compare the results obtained using the sub-hypothesis decomposition method with those obtained without it. The results without hypothesis decomposition are presented at the top of the table, while those with hypothesis decomposition are shown below. Benchmark ARC-ID37.}
    \centering
    \scalebox{0.8}{
    \begin{tabular}{ccc}
\toprule
\textbf{Observations} & \textbf{Rounds} & \textbf{Executable Function} \\ \midrule
\multirow{2}{*}{\thead{\includegraphics[width=4cm]{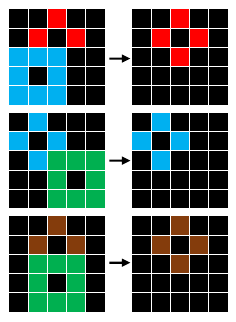}}} & \thead{\textbf{Round 1:}\\Step 1: Identify the third row\\Step 2: Check for a non-zero \\ numbers in the third row.\\ Step 3: Replace the number at \\the center position of the sequence. \\ Step 4: Change all numbers \\in rows 4 and 5 to zero.} & \thead{
\begin{lstlisting}[language=Python, style=mystyle]
def fn(grid):
  # Step 1: Identify the third row
  third_row = grid[2]
  # Step 2: Check for a sequence of three or more identical non-zero numbers
  n = len(third_row)
  i = 0
  while i < n:
    if third_row[i] != 0:
      # Find the length of the sequence of identical numbers
      start = i
      while i < n and third_row[i] == third_row[start]:
        i += 1
      length = i - start
      # If the sequence is three or more, replace it
      if length >= 3:
        # Step 3: Replace the sequence with the number from the second row
        center_index = start + length // 2
        replacement_number = grid[1][center_index]
        third_row[start:i] = [0] * n  # Set the entire row to zero first
        third_row[center_index] = replacement_number
        break  # Assuming only one sequence needs to be replaced
    else:
      i += 1
  # Step 4: Set all numbers in rows 4 and 5 to zero
  grid[3] = [0] * n
  grid[4] = [0] * n
  return grid
\end{lstlisting}} \\ \cline{2-3} 
& \thead{\textbf{Round 2:}\\Step 1: Identify the third row\\Step 2: Check for a non-zero\\ numbers in the third row.\\ Step 3: Replace identical numbers\\ in the third row with the\\ corresponding non-zero  number\\ from the first row.} & \thead{
\begin{lstlisting}[language=Python, style=mystyle]
def fn(grid):
  # Step 1: Identify the third row
  third_row = grid[2]   
  # Step 2: Check for a sequence of three or more identical non-zero numbers
  n = len(third_row)
  i = 0
  while i < n:
    if third_row[i] != 0:
      # Find the length of the sequence of identical numbers
      start = i
      while i < n and third_row[i] == third_row[start]:
        i += 1
      length = i - start            
      # If the sequence is three or more, replace it
      if length >= 3:
        # Step 3: Replace the sequence
        # Find the number from the first row at the same column
        replacement_number = grid[0][start]
        # Replace the sequence with the replacement number
        third_row[start:i] = [replacement_number] + [0] * (length - 1)
    else:
      i += 1
  # Step 4: Change all numbers in rows 4 and 5 to zero
  for row in range(3, 5):
    grid[row] = [0] * n
  return grid
\end{lstlisting}} \\
\bottomrule
\end{tabular}
    }
    \label{tab:case_study_feed2}
\vspace{-4mm}
\end{table*}

\begin{table*}[ht!]
    \caption{We compare the results obtained using the sub-hypothesis decomposition method with those obtained without it. The results without hypothesis decomposition are presented at the top of the table, while those with hypothesis decomposition are shown below. Benchmark List Function-ID2.}
    \centering
    \scalebox{0.8}{
    \begin{tabular}{ccc}
\toprule
\textbf{Observations} & \textbf{Hypothesis} & \textbf{Executable Function} \\ \midrule
\multirow{2}{*}{\thead{\scriptsize{\texttt{[]}} $\rightarrow$ \scriptsize{{\texttt{[]}}} \\ \scriptsize{\texttt{[6, 9]}} $\rightarrow$ \scriptsize{{\texttt{[]}}} \\ \scriptsize{\texttt{[1, 5, 0, 6, 2, 9, 3]}} $\rightarrow$ \scriptsize{\texttt{[0]}}  \\ \scriptsize{\texttt{[6, 3, 4, 1, 7, 2, 9, 8, 0]}} $\rightarrow$ \scriptsize{\texttt{[4]}} \\ \dots }} & \thead{\textbf{No Sub Hypothesis:}\\The output is the element \\ from the input list that \\ is exactly in the middle \\ of the list.} & \thead{
\begin{lstlisting}[language=Python, style=mystyle]
def fn(input_list):
  # Check if the list is empty or has only one element
  if len(input_list) <= 1:
    return []
  # Calculate the middle index
  middle_index = (len(input_list) - 1) // 2
  # Return the middle element as a list
  return [input_list[middle_index]]
\end{lstlisting}} \\ \cline{2-3} 
& \thead{\textbf{Sub Hypothesis:}\\Step 1: If the list has fewer \\ than three elements,\\ the output is an empty list.\\ Step 2:If the list has three \\ or more elements,\\ the output is a list containing 
 \\ only the integer at the third \\ position.} & \thead{
\begin{lstlisting}[language=Python, style=mystyle]
def fn(input_list):
  # Check if the list has fewer than three elements
  if len(input_list) < 3:
    return []
  else:
  # Return a list containing the integer
  # at the third position (index 2)
    return [input_list[2]]
    
\end{lstlisting}} \\
\bottomrule
\end{tabular}
    }
    \label{tab:case_study_hyp2}
\vspace{-4mm}
\end{table*}

\begin{table*}[ht!]
    \caption{We compare the results obtained using the sub-hypothesis decomposition method with those obtained without it. The results without hypothesis decomposition are presented at the top of the table, while those with hypothesis decomposition are shown below. Benchmark Livecodebench Input-ID37.}
    \centering
    \scalebox{0.8}{
    \begin{tabular}{ccc}
\toprule
\textbf{Observations} & \textbf{Hypothesis} & \textbf{Executable Function} \\ \midrule
\multirow{2}{*}{\thead{
\begin{lstlisting}[language=Python, style=mystyle]
def minOperations(a, b):
  def f():
    ret = 0
    aa, bb = a[:], b[:]
    for i in range(len(a)):
      if a[i] > a[-1] or b[i] > b[-1]:
        a[i], b[i] = b[i], a[i]
        ret += 1
      if a[i] > a[-1] or b[i] > b[-1]:
        return inf
    a[:] = aa
    b[:] = bb
    return ret
  
    ans = f()
    a[-1], b[-1] = b[-1], a[-1]
    ans = min(ans, f() + 1)
    return -1 if ans > len(a) else ans
    # assert f(??) == 1
\end{lstlisting}}} & \thead{\textbf{No Sub Hypothesis:}\\The function `minOperations' attempts \\to determine the minimum number \\of swaps needed to ensure \\that for each index `i',\\ neither `a[i]' nor `b[i]'\\ are greater than the last\\ elements of their respective lists.} & assert minOperations([3, 1, 2], [1, 3, 2]) == 1 \\ \cline{2-3} 
& \thead{\textbf{Sub Hypothesis:}\\Step 1: \textbf{Function Purpose} \\The goal of `minOperations` is to ... \\ Step 2: \textbf{Inner Function} \\ It attempts to ensure \\ that for each `i', ... \\ Step 3: \textbf{Main Logic} \\ It calculates `ans` by calling `f', ... \\ Step 4: \textbf{Objective} \\ To find an input such that... } & \thead{assert minOperations([2, 3], [3, 2]) == 1} \\
\bottomrule
\end{tabular}
    }
    \label{tab:case_study_hyp3}
\vspace{-4mm}
\end{table*}

\begin{table*}[ht!]
    \caption{We present the additional improvements resulting from the use of amendments. The results shown above are those obtained after the initial hypotheses, while the results displayed below reflect the outcomes following the submission of amendments and subsequent reflections. Benchmark CruxEval Output-ID328.}
    \centering
    \scalebox{0.8}{
    \begin{tabular}{ccc}
\toprule
\textbf{Observations} & \textbf{Rounds} & \textbf{Executable Function} \\ \midrule
\multirow{2}{*}{\thead{
\begin{lstlisting}[language=Python, style=mystyle]
def f(array, L):
  if L <= 0:
    return array
  if len(array) < L:
    array.extend(f(array, L - len(array)))
  return array
# assert f([1, 2, 3], 4) == ??
\end{lstlisting}}} & \thead{\textbf{Round 1:} \\ Step 1 Base Case Check: \\The function checks if `L' is\\ less than or equal to 0.\\ If true, it simply returns\\ the array as is \\ Step 2 Recursive Extension: \\If the length of `array' \\is less than L, it then\\ calls itself recursively \\ Step 3 Return Array:\\If the array is already of\\ length `L` or longer, the\\ function simply returns the \\array without any modifications.} & assert f([1, 2, 3], 4) == [1, 2, 3, 1] \\ \cline{2-3} 
& \thead{\textbf{Round 2:}\\ Step 1 Base Case Check: ...\\ Step 2 Recursive Extension: ...\\ Step 3 Return Array: ...\\ Step 4 Example Check: \\The original call then extends\\ `[1, 2, 3]' by `[1, 2, 3]',\\ resulting in `[1, 2, 3, 1, 2, 3]'.} & \thead{assert f([1, 2, 3], 4) == [1, 2, 3, 1, 2, 3]} \\
\bottomrule
\end{tabular}
    }
    \label{tab:case_study_feed3}
\vspace{-4mm}
\end{table*}

We validated the effectiveness of the proposed method using examples from various benchmarks. For instance, as shown in Table~\ref{tab:case_study_feed2}, the MiniARC task example with ID 37 highlights how the LLM, after receiving feedback, successfully reflects on its errors and submits a revised solution.

In Table~\ref{tab:case_study_hyp2}, hypothesis decomposition reveals that the output number is determined not only by its position at the middle of the input array but also by being the third character.

In Table~\ref{tab:case_study_hyp3}, compared to models without hypothesis decomposition, those utilizing this approach progressively analyze the function's behavior, ultimately achieving an abstract understanding of the program and making accurate assertions. In Table~\ref{tab:case_study_feed3}, for a complex recursive function, while the LLM accurately grasped the overall functionality of the function through hypothesis decomposition, it encountered difficulties during the detailed analysis of specific instances. Following the submission of a revised solution, the LLM reflected on its errors and successfully resolved the issue, addressing the collapse of the overall logical chain caused by a failure in a single step during multi-step reasoning.

\subsection{Failure Analyze}
\label{app:failure}

\begin{table*}[ht!]
    \caption{Failure case study. The observations indicate that the hypotheses proposed by the LLM are overly simplistic, rendering it difficult to account for all possible cases. Additionally, manual inspection and efforts to summarize these hypotheses proved challenging. This limitation is one of the primary factors contributing to the LLM's failure in this task. Benchmark: List Function-ID9.}
    \centering
    \scalebox{0.8}{
    \begin{tabular}{ccc}
\toprule
\textbf{Observations} & \textbf{Rounds} & \textbf{Executable Function} \\ \midrule
\multirow{2}{*}{\thead{\scriptsize{\texttt{[2, 8]}} $\rightarrow$ \scriptsize{{\texttt{[8]}}} \\ \scriptsize{\texttt{[7, 5, ..., 8, 4]}} $\rightarrow$ \scriptsize{\texttt{[5, ..., 8]}} \\ \scriptsize{\texttt{[8, 2, ..., 9]}} $\rightarrow$ \scriptsize{\texttt{[2, ..., 9]}} \\ \scriptsize{\texttt{[3, 2, 1, 0, 7, 8]}} $\rightarrow$ \scriptsize{\texttt{[2, 1, 0]}} \\ \dots \\ \dots }} &
\thead{\textbf{Round 1:} \\ Remove the first element \\ from the input list.} & \thead{
\begin{lstlisting}[language=Python, style=mystyle]
def fn(input_list):
  """
  Removes the first element from the
  input list of integers.
  """
  # Check if the input list is empty
  if not input_list:
    return []

  # Return the list excluding the
  # first element
  return input_list[1:]
\end{lstlisting}} \\ \cline{2-3} & \thead{\textbf{Round 2:} \\Remove the first and the last \\elements from the input list.} & \thead{
\begin{lstlisting}[language=Python, style=mystyle]
def fn(input_list):
  """
  Remove the first and last elements
  from the input list.
  """
  # Check if the list has less than
  # 2 elements
  if len(input_list) < 2:
    return []  
  # Return the list excluding the
  # first and last elements
  return input_list[1:-1]
\end{lstlisting}} \\
 \bottomrule
\end{tabular}
    }
    \label{tab:fail1}
\vspace{-4mm}
\end{table*}

\begin{table*}[ht!]
    \caption{Failure case study. The LLM make some redundant hypotheses, which led to failure. Even after amendments, it was unable to break free from its original thought framework, resulting in a failure to reflect effectively. Benchmark: ARC-ID5.}
    \centering
    \scalebox{0.8}{
    \begin{tabular}{ccc}
\toprule
\textbf{Observations} & \textbf{Rounds} & \textbf{Executable Function} \\ \midrule
\multirow{2}{*}{\thead{\includegraphics[width=3.5cm]{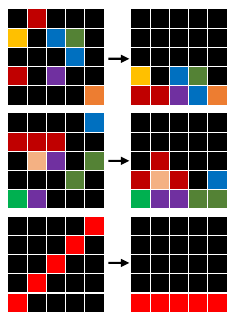}}} &
\thead{\textbf{Round 1:} \\ Step 1: Identify non-zero elements\\ in the input grid. \\ Step 2: Shift all non-zero elements \\ down one row. \\ Step 3: Repeat the process for each \\ input-output pair to verify \\ consistency.} & \thead{
\begin{lstlisting}[language=Python, style=mystyle]
def fn(grid):
  if not grid or not grid[0]:
    return grid
  num_rows = len(grid)
  num_cols = len(grid[0])
  # Create a new grid initialized with zeros
  new_grid = [[0] * num_cols for _ in range(num_rows)]
  for col in range(num_cols):
    # Collect non-zero elements in the current column
    non_zero_elements = [grid[row][col] for row in range(num_rows) \\
    if grid[row][col] != 0]
    # If there are non-zero elements, shift them down by one row
    if non_zero_elements:
      # Shift down by one row with wrap-around
      shifted_elements = [0] * num_rows
      for i, elem in enumerate(non_zero_elements):
        new_row = (i + 1) % num_rows
        shifted_elements[new_row] = elem
      # Find the maximum value among the shifted elements
      max_value = max(shifted_elements)

      # Fill the entire column in the new grid with the maximum value
      for row in range(num_rows):
        new_grid[row][col] = max_value
  return new_grid
\end{lstlisting}} \\ \cline{2-3} & \thead{\textbf{Round 2:} \\ Step 1:Identify non-zero elements\\ in the input grid. \\ Step 2: Shift all non-zero\\ elements down one row. \\ Step 3: In the resulting grid, fill each\\ row with the maximum value\\ from its respective column,\\ considering only the shifted \\non-zero elements.\\ Step 4: Repeat the process for each \\ input-output pair to verify consistency.} & \thead{
\begin{lstlisting}[language=Python, style=mystyle]
def fn(grid):
  if not grid or not grid[0]:
    return grid    
  rows, cols = len(grid), len(grid[0])    
  # Step 1: Identify non-zero elements and their positions
  non_zero_positions = []
  for r in range(rows):
    for c in range(cols):
      if grid[r][c] != 0:
        non_zero_positions.append((r, c))
    
  # Step 2: Shift non-zero elements down by one row
  shifted_positions = [( (r + 1) % rows, c) for r, c in \\
  non_zero_positions]
    
  # Step 3: Determine the maximum value for each column 
  # from the shifted positions
  max_values = [0] * cols
  for r, c in shifted_positions:
    max_values[c] = max(max_values[c], grid[r][c])
    
  # Step 4: Construct the new grid
  new_grid = [[0] * cols for _ in range(rows)]
  for r, c in shifted_positions:
    new_grid[r][c] = max_values[c]
    
  return new_grid
\end{lstlisting}} \\
 \bottomrule
\end{tabular}
    }
    \label{tab:fail2}
\vspace{-4mm}
\end{table*}

We analyze RHDA's performance in numerous failure cases and summarize the underlying causes of these failures. Our findings suggest that the primary reason can be attributed to the insufficient intrinsic capability of LLMs in code reasoning tasks. This limitation is specifically reflected in two aspects:  
\begin{itemize}
    \item \textbf{Sub-hypotheses fail to address the problem}: For tasks that are overly complex or abstract (e.g., cases shown in Table~\ref{tab:fail1}), even though hypothesis decomposition attempts to reduce task complexity, LLMs still struggle to handle them effectively.  
    \item \textbf{Amendments fail to correct sub-hypotheses}: While amendments leverage external feedback to help LLMs reflect on their mistakes, the models often remain confined to their existing thought framework, even after recognizing errors (e.g., cases shown in Table~\ref{tab:fail2}). This results in the correction failing to resolve the issue.  
\end{itemize}
These observations indicate that for tasks exceeding the intrinsic capabilities of LLMs, relying solely on reflective hypothesis decomposition and amendment may not be sufficient to improve the model's performance.

\section{Costs}
\label{app:costs}
\begin{table}[!ht]
\centering
\caption{Avg. API calls and Total Cost using GPT-4o.}
\scalebox{0.65}{
\begin{tabular}{lcccccccc}
\toprule
\multirow{2}{*}{Method} & \multicolumn{4}{c}{Avg. API Calls}                            & \multicolumn{4}{c}{Total Cost (cent)}                         \\ \cmidrule(lr){2-5} \cmidrule(lr){6-9} 
 &
  \multicolumn{1}{l}{List Func} &
  \multicolumn{1}{l}{MiniARC} &
  \multicolumn{1}{l}{RobustFill} &
  \multicolumn{1}{l}{Deepcoder} &
  \multicolumn{1}{l}{List Func} &
  \multicolumn{1}{l}{MiniARC} &
  \multicolumn{1}{l}{RobustFill} &
  \multicolumn{1}{l}{Deepcoder} \\ \midrule
IO                               & 8.0           & 4.0           & 5.0            & 3.0          & 10.2           & 4.6          & 2.0            & 3.3          \\
PoT                              & 1.0           & 1.0           & 1.0            & 1.0          & 5.0            & 3.7          & 0.6            & 1.2          \\
CoC                              & 1.0           & 1.0           & 1.0            & 1.0          & 11.0           & 9.0          & 1.1            & 1.4          \\
SC (N=3)                         & 3.0           & 24.0          & 15.0           & 9.0          & 5.3            & 3.7          & 0.6            & 1.2          \\
SR (T=2)                         & 1.4           & 1.9           & 1.5            & 1.6          & 4.6            & 3.3          & 0.5            & 1.1          \\
T=2, N=3                         & 5.4           & 5.9           & 5.5            & 5.6          & 8.6            & 4.0          & 3.1            & 4.7          \\ \midrule
\multirow{2}{*}{Method} & \multicolumn{2}{c}{Deductive} & \multicolumn{2}{c}{Abductive} & \multicolumn{2}{c}{Deductive} & \multicolumn{2}{c}{Abductive} \\ \cmidrule(lr){2-3} \cmidrule(lr){4-5} \cmidrule(lr){6-7} \cmidrule(lr){8-9}
 &
  \multicolumn{1}{l}{CRUXEval} &
  \multicolumn{1}{l}{LiveCodeBench} &
  \multicolumn{1}{l}{CRUXEval} &
  \multicolumn{1}{l}{LiveCodeBench} &
  \multicolumn{1}{l}{CRUXEval} &
  \multicolumn{1}{l}{LiveCodeBench} &
  \multicolumn{1}{l}{CRUXEval} &
  \multicolumn{1}{l}{LiveCodeBench} \\ \midrule
Standard                         & 1.0           & 1.0           & 1.0            & 1.0          & 2.9            & 0.5          & 3.1            & 1.5          \\
CoT                              & 1.0           & 1.0           & 1.0            & 1.0          & 19.4           & 3.7          & 19.5           & 3.3          \\
SC (N=3)                         & 3.0           & 3.0           & 3.0            & 3.0          & 2.9            & 0.5          & 3.3            & 1.5          \\
SR (T=2)                         & 1.6           & 1.7           & 1.4            & 1.5          & 3.8            & 0.6          & 3.4            & 1.6          \\
CoC                              & 1.0           & 1.0           & 1.0            & 1.0          & 18.3           & 4.1          & 19.0           & 3.4          \\
T=2, N=1                         & 1.6           & 1.7           & 1.4            & 1.5          & 19.0           & 4.4          & 18.8           & 4.4          \\ \bottomrule
\end{tabular}}
\label{tab:cost}
\end{table}

In Table~\ref{tab:cost}, we present the average number of API calls and the total cost for each task. We used GPT-4o, with an input cost of \$0.0025/1K tokens and an output cost of \$0.01/1K tokens. The results indicate that our approach still demonstrates high cost-effectiveness for certain tasks.

\section{Trade off between number of iterations and performance gain}
\begin{figure}[ht!]
    \centering
    \includegraphics[width=0.85\textwidth]{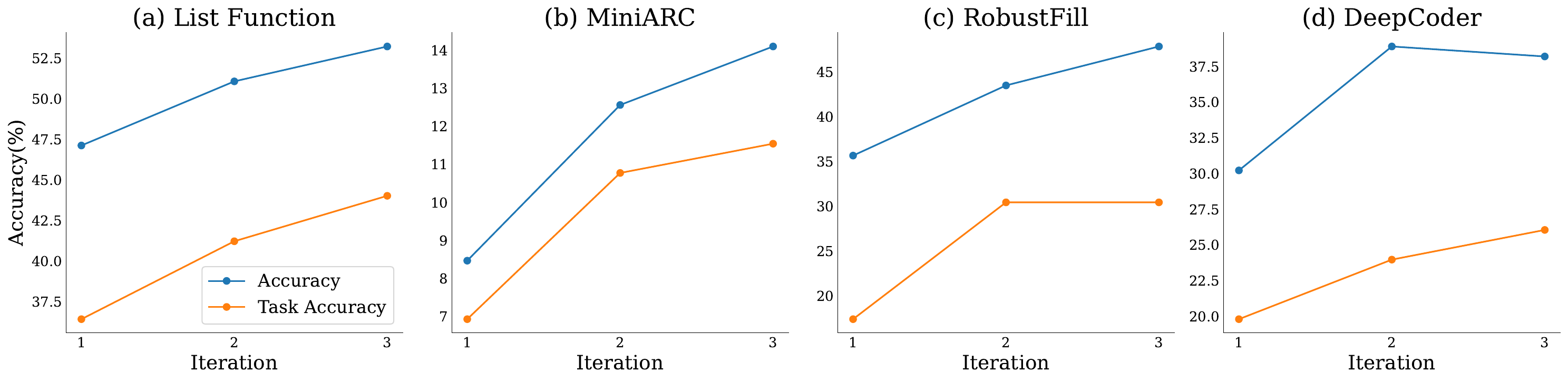}
    \caption{In the inductive code reasoning tasks, as the number of iterations increased, the performance continued to improve.}
    \label{fig:iter1}
\end{figure}
\begin{figure}[ht!]
    \centering
    \includegraphics[width=0.5\textwidth]{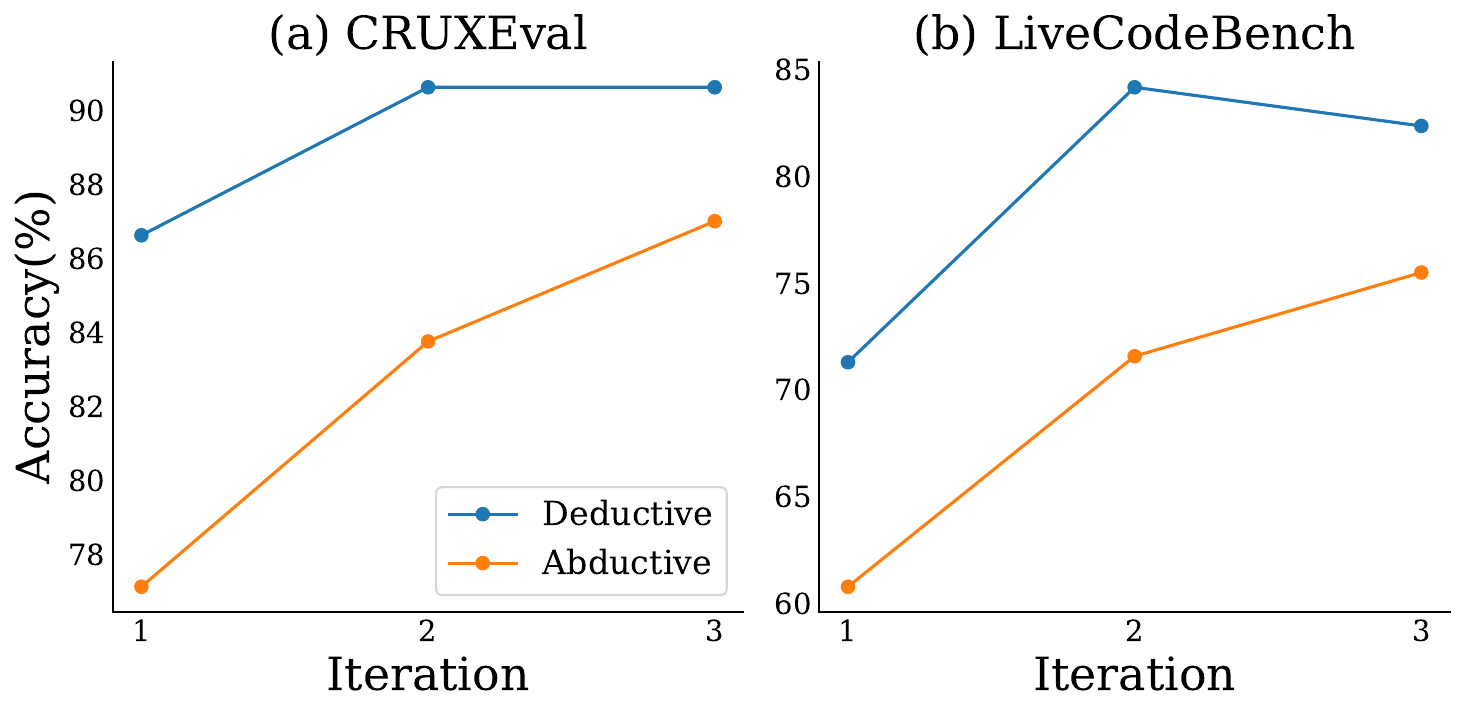}
    \caption{In the deductive code reasoning tasks, the performance slightly decreased as the number of iterations increased. Conversely, in the abductive code reasoning tasks, the performance consistently improved with an increasing number of iterations.}
    \label{fig:iter2}
\end{figure}
In this section, we investigate the impact of iteration count on the performance of three types of reasoning tasks, with experimental results illustrated in Figure~\ref{fig:iter1} and Figure~\ref{fig:iter2}. For inductive and abductive code reasoning tasks, performance consistently improved as the number of iterations increased. However, the rate of improvement diminished, with marginal gains becoming less significant at higher iteration counts. Conversely, for deductive code reasoning tasks, performance followed a rise-and-fall trend, initially improving but declining with excessive iterations. These findings suggest that while increasing the number of iterations can enhance performance for general code reasoning tasks, it is crucial to balance iterative gains against potential performance instability.
\section{Prompts}
\label{app:prompts}
\begin{table*}[ht!]
    \centering
    \caption{Prompts used in our study. \{\} refers to a placeholder.}
    \scalebox{0.9}{
    \begin{tabular}{ll}
        \toprule
        Type & Prompt \\
        \midrule
        \thead{Sub Hypothesis \\ Generation} & \texttt{\thead{Generate a rule that maps the following inputs to their \\ corresponding outputs step by steps. \textbf{\{Task description\}} \\\\ \textbf{\{Examples\}} \\\\ Please format your rule as follows: \\\\  \textbf{\{Rule format\}}}}\\
        \midrule
        \thead{Amendment \\ Submission} & \texttt{\thead{Your rule: \textbf{\{Rule\}} \\\\ This rule does not work for the following examples. \\\\\textbf{\{Feedback\}} \\\\ Please carefully reconsider each of your steps to ensure \\that the rules are correct. Systematically\\ generate new rules, step by step.\\ \textbf{\{Feedback description\}} Please \\ format your rule as follows: \\\\  \textbf{\{Rule format\}}}}\\
        \midrule
        \thead{Hypothesis \\ Translation} & \texttt{\thead{You are an expert Python programmer. Write a Python \\ function `fn` for the following rule. \textbf{\{Translation} \\ \textbf{Example description\}}\\\\ Rule: \textbf{\{Rule\}}}}\\
        \midrule
        \thead{Rule \\ Application} & \texttt{\thead{Generate an output corresponding to the given input based \\ on the rule. \textbf{\{Application Example description\}}\\\\ Rule: \textbf{\{Rule\}} \\\\ Input: \textbf{\{Test input\}} \\Output:}}\\
        \bottomrule
    \end{tabular}
    }
    \label{tab:prompts}
\end{table*}

\end{document}